\documentclass[fleqn,12pt]{wlscirep}
\usepackage[utf8]{inputenc}
\usepackage[T1]{fontenc}
\usepackage{mathptmx}
\usepackage{bbding}
\usepackage{hyperref}
\usepackage{graphicx}  
\usepackage{float}  
\usepackage{subfigure}  
\usepackage{multirow}
\usepackage{ragged2e}
\usepackage[normalem]{ulem}
\useunder{\uline}{\ul}{}
\usepackage{multibib}
\newcites{methods}{Method}

\usepackage{placeins}
\usepackage[framemethod=TikZ]{mdframed}
\usepackage{adjustbox}
\mdfsetup{
  roundcorner=10pt,
  linewidth=1pt,
  innertopmargin=10pt,
  innerbottommargin=10pt,
  innerrightmargin=10pt,
  innerleftmargin=10pt,
  backgroundcolor=gray!10
}
\usepackage{listings}
\usepackage{xcolor}
\usepackage{svg}

\lstdefinelanguage{json}{
    basicstyle=\footnotesize\ttfamily,
    numberstyle=\scriptsize,
    stepnumber=1,
    numbersep=8pt,
    showstringspaces=false,
    breaklines=true,
    frame=none, 
    backgroundcolor=\color{white}, 
    literate=
     *{:}{{{\color{purple}{:}}}}{1}
      {,}{{{\color{purple}{,}}}}{1}
      {\{}{{{\color{blue}{\{}}}}{1}
      {\}}{{{\color{blue}{\}}}}}{1}
      {[}{{{\color{blue}{[}}}}{1}
      {]}{{{\color{blue}{]}}}}{1},
}

\title{Agent Hospital: A Simulacrum of Hospital with Evolvable Medical Agents}

\author[1, 2]{Junkai Li}
\author[1, 2]{Yunghwei Lai}
\author[1, 2]{Weitao Li}
\author[1, 2]{Jingyi Ren}
\author[1]{Meng Zhang}
\author[1, 2]{Xinhui Kang}
\author[1]{Siyu Wang}
\author[1]{Peng Li}
\author[1]{Ya-Qin Zhang}
\author[1\Envelope]{Weizhi Ma}
\author[1, 2 \Envelope]{Yang Liu}
\affil[1]{Institute for AI Industry Research (AIR), Tsinghua University, China}
\affil[2]{Department of Computer Science and Technology, Tsinghua University, China}
\affil[\Envelope]{E-mail: \tt mawz@tsinghua.edu.cn; liuyang2011@tsinghua.edu.cn}

\begin{abstract}
The recent rapid development of large language models (LLMs) \cite{OpenAI:24:GPT4,Meta:23:Llama} has sparked a new wave of technological revolution in medical artificial intelligence (AI) \cite{Moor:23:GMAI,Singhal:23:Med-PaLM}.
While LLMs are designed to understand and generate text like a human, autonomous agents that utilize LLMs as their ``brain'' have exhibited capabilities beyond text processing such as planning, reflection, and using tools by enabling their ``bodies'' to interact with the environment \cite{Park:23:Smallville,Yao:23:ReAct,Schick:23:Toolformer}. 
We introduce a simulacrum of hospital called \textit{Agent Hospital} that simulates the entire process of treating illness, in which all patients, nurses, and doctors are LLM-powered autonomous agents. Within the simulacrum, doctor agents are able to evolve by treating a large number of patient agents without the need to label training data manually.
After treating tens of thousands of patient agents in the simulacrum (human doctors may take several years in the real world), the evolved doctor agents outperform state-of-the-art medical agent methods on the MedQA benchmark \cite{Jin:21:MedQA} comprising US Medical Licensing Examination (USMLE) test questions. 
Our methods of simulacrum construction and agent evolution have the potential in benefiting a broad range of applications beyond medical AI.
\end{abstract}

\begin{document}
\flushbottom
\maketitle

\section*{Introduction}

\begin{figure*}[!t]
    \centering
    \includegraphics[width=\textwidth]{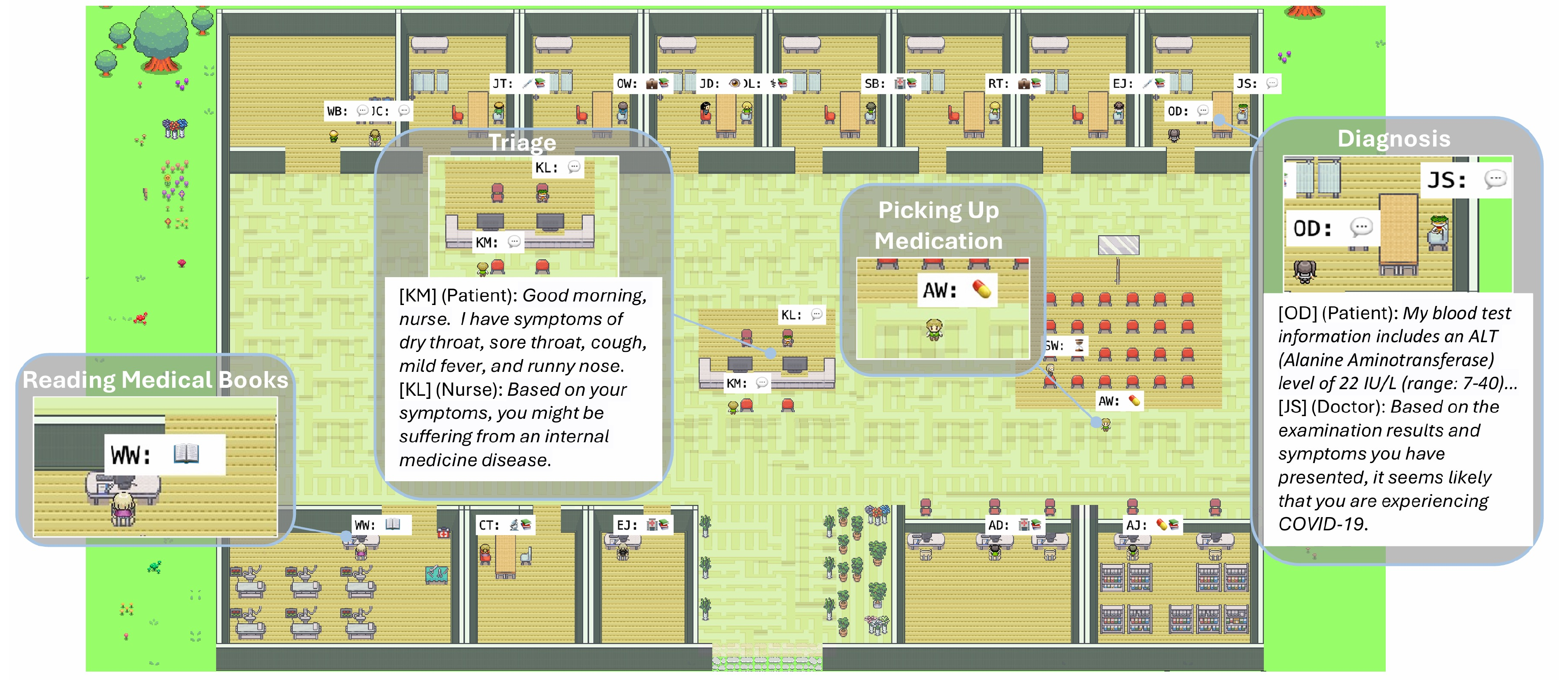}
    \caption{\textbf{An overview of Agent Hospital}. {\it Agent Hospital} is a simulacrum of hospital in which patients, nurses, and doctors are autonomous agents powered by large language models. {\it Agent Hospital} simulates the whole closed cycle of treating a patient's illness: disease onset, triage, registration, consultation, medical examination, diagnosis, medicine dispensary, convalescence, and post-hospital follow-up visit. Doctor agents can keep improving treatment performance over time by reading medical textbooks and treating patient agents. An interesting finding is that the expertise doctor agents acquired in the virtual world is applicable to solving real-world medicare problems.}
\label{fig:overview}
\end{figure*}

Becoming a medical professional is a long, hard haul. 
It often takes a medical student 12 years at school, four years at college, and four years at medical school to complete general education requirements and take medical courses. 
After 20 years of school, there is still much for the medical student to learn at hospital during three years of residency before finally becoming an attending physician. 
As a result, the path to becoming a doctor can be roughly divided into two phases: (1) acquiring knowledge from textbooks at school and (2) acquiring expertise from practice at hospital.

Most recent advances in medical artificial intelligence (AI) \cite{Moor:23:GMAI,Singhal:23:Med-PaLM,Li:23:Llava-med, Moor:23:Med-Flamingo,Tu:24:MedPaLM_Med,Ma:24:GHAI} have concentrated on the first phase (i.e., medical knowledge acquisition) by training large language models (LLMs) \cite{OpenAI:24:GPT4,Meta:23:Llama,Liu:23:Llava} tailored for medicine on enormous amounts of textual data. 
For example, the training corpus of Med-PaLM \cite{Singhal:23:Med-PaLM}, which is a 540-billion parameter LLM, contains 780 billion tokens representing a mixture of webpages, Wikipedia articles, source code, social media conversations, news articles, books, and medical textbooks. After acquiring general and medical knowledge from massively large data, Med-PaLM is reported to be the first LLM to reach the human expert level on answering the US Medical Licensing Examination (USMLE) style questions. 
Despite the success of medical LLMs, it is hard to directly use them to model medical expertise acquisition at hospital because they are designed to provide foundational capabilities of understanding and generating human languages rather than dealing with task-specific scenarios in the real world.

The rise of LLM-powered autonomous agents \cite{Park:23:Smallville,Yao:23:ReAct,Schick:23:Toolformer,Wei:22:CoT,Yao:23:ToT,Shinn:23:Reflexion,Zelikman:22:Star,Shen:23:Hugginggpt} brings hope to modeling the second phase (i.e., medical expertise acquisition). 
Compared with LLMs, autonomous agents are more like humans. Using LLMs as their ``brain'' to think, agents are able to act in an environment with their ``bodies'' autonomously\cite{Wang:23:Voyager,Gao:23:S3,Hua:23:WarAgent, Wang:23:RecAgent, Zhang:24:Generative_recagent,Williams:23:epidemic_generative_agent, Xiao:23:simulating_public, Xu:23:Werewolf,Zhao:23:Competeai,Li:TradingGPT,Hong:23:MetaGPT,Qian:23:ChatDev,Guo:24:multiagent_survey,Chen:23:Agentverse, Yang:24:UA3,Liu:24:DyLAN,Yang:24:ActRe}. 
For example, Smallville \cite{Park:23:Smallville}, which is a sandbox game world where 25 agents live and work, has demonstrated that human behaviors can be simulated by agents. These agents are able to plan their days, go to work, chat with neighbors, and reflect on days past. After interacting with each other continuously over two full game days in Smallville, they produce emergent social behaviors such as sharing news, forming relationships, and coordinating group activities. 
While current research on medical agents has focused on multi-agent collaboration for medical reasoning \cite{Tang:23:MedAgents,Fan:24:AI_Hospital,Li:24:Mmedagent,Kim:24:MDAgents,Wei:24:MedAide,Kim:24:colla_llms_medical,Lievin:24:medical_reasoning,Nori:23:Medprompt}, how to enable doctor agents to acquire medical expertise from practice like humans do at hospital still remains a challenge.

In this work, we introduce a simulacrum of hospital called {\it Agent Hospital} to simulate medical expertise acquisition. 
As shown in Figure \ref{fig:overview}, {\it Agent Hospital} is a virtual world in which all patients, nurses, and doctors are LLM-powered autonomous agents. It functions like a real-world hospital. 
Patient agents will go to {\it Agent Hospital} if they get sick. At the triage station, nurse agents ask patient agents about their symptoms. Then, following the nurse's suggestions, patient agents go to the registration desk, wait to consult doctor agents, have medical examinations, get diagnosis results, pick up medication, and go back home. If patient agents recover after several days, they will express their gratitude to nurse and doctor agents. Otherwise, they will go to {\it Agent Hospital} again, complain to nurse and doctor agents, and start another round of treatment cycle. 
If a doctor agent has successfully treated a patient agent, the case will be recorded to offer a reference for future treatment. The doctor agent can also benefit from failure by reflecting to gain experience to avoid making the same mistake in the future \cite{Yang:23:Rule_Accumulation}. Besides caring for patient agents, doctor agents also read medical books in their spare time to consolidate knowledge and expertise. 
As time in {\it Agent Hospital} passes several magnitude orders faster than in the real world, the number of patent agents that a doctor agent can treat is accordingly much higher than a human doctor does during lifespan. Therefore, doctor agents can evolve over a long time span in {\it Agent Hospital} and keep improving medical proficiency similar to AlphaGo Zero \cite{Silver:17:AlphaGo-Zero}.

The AI technique behind {\it Agent Hospital} is a new paradigm named {\it Simulacrum-based Evolutionary Agent Learning} (SEAL). SEAL consists of two components: {\it simulacrum construction} and {\it agent evolution}. 
Similar to establishing a world model \cite{LeCun:22:WorldModel}, simulacrum construction aims to build a simulacrum of hospital capable of generating a large amount of medical data for doctor agents to acquire medical expertise: the disease that a patient agent suffers from, the symptoms that a patient agent experiences, the result of clinical examination, and disease progression after the patient agent follows the doctor agent's prescribed treatment plan. To do so, we propose to couple LLMs with medical knowledge bases in a flexible way: LLMs generate medical data guided by medical knowledge bases. Therefore, all the training data is generated by the virtual world rather than being annotated by humans. 
After simulacrum construction, agent evolution aims to enable doctor agents to keep acquiring medical expertise from both successful and unsuccessful cases of treatment over time. This can be done by storing and retrieving successful cases for reference and gaining experience from unsuccessful cases.

We evaluated our approach in both virtual and real worlds. 
In the virtual world, the proficiency of doctor agents is assessed on three tasks: {\it medical examination selection} (whether a doctor agent makes a correct decision on medical examination), {\it diagnosis} (whether a doctor agent identifies the disease correctly), and {\it treatment plan recommendation} (whether a doctor agent recommends a correct treatment plan). {\it Agent Hospital} comprises 32 departments that cover 339 diseases (details are provided in Appendix A.1 \& A.2). In the beginning, doctor agents can only use general and medical knowledge encoded in an LLM. In {\it Agent Hospital}, doctor agents evolve by treating patient agents and reading textbooks. We find that the diagnostic accuracy of doctor agents keeps improving with the increase of the number of patient agents being treated, suggesting that doctor agents seem to acquire medical expertise from practice in {\it Agent Hospital}. 
Interestingly, the medical skills that doctor agents learned in {\it Agent Hospital} are applicable to the real world. We observe that the accuracy of doctor agents answering questions in the MedQA dataset\cite{Jin:21:MedQA} also improves with the increase of the number of patient agents being treated. Thanks to the scaling laws of evolution, evolved doctor agents outperform existing methods on the MedQA dataset without using labeled training data of the benchmark.

The main contribution of our work is to propose a new framework for solving task-specific problems in real-world scenarios. 
Instead of tailoring LLMs to a specific use case and annotating data manually, SEAL advocates building a simulacrum according to the workflow of the use case and generating data automatically. This not only directly accommodates the requirements of specific applications, but also significantly reduces the overhead for labeling data. Another benefit that SEAL brings to vertical applications is eliminating the need for training domain-specific LLMs. As shown in {\it Agent Hospital}, SEAL couples foundation models with domain knowledge bases in a flexible way, which are both readily available and plug-and-play. 
Therefore, we believe that SEAL has the potential to be applied to a broad range of applications beyond medical AI in the future.

\begin{figure*}[!t]
    \centering
    \includegraphics[width=0.9\textwidth]{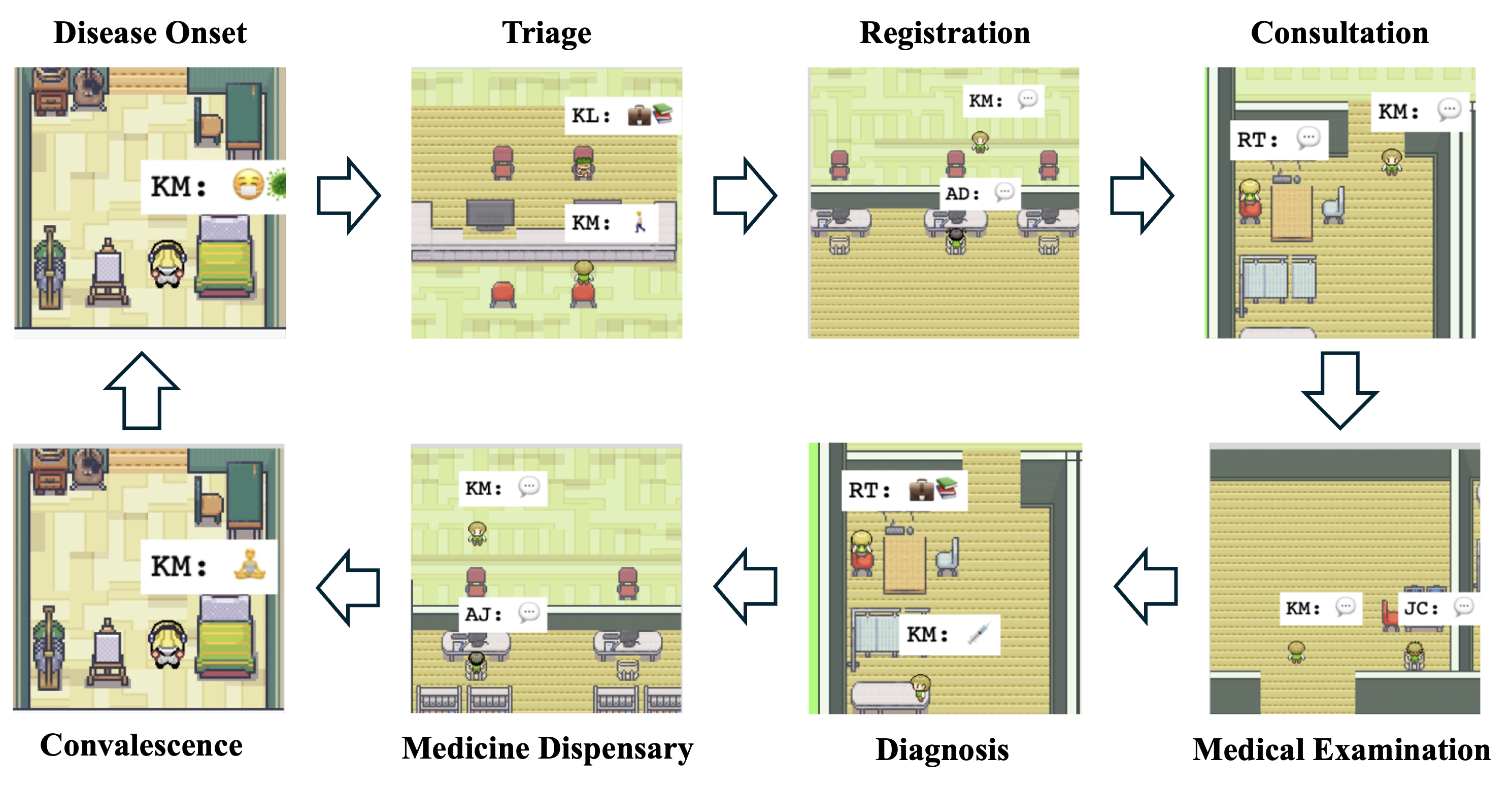}
    \caption{\textbf{Agent Hospital simulates the whole closed cycle of treating illness}. In this example, patient agent Kenneth Morgan falls ill and visits {\it Agent Hospital}. Triage nurse Katherine Li conducts an initial evaluation of Mr. Morgan's symptoms and refers him to the dermatology department. Mr. Morgan then registers at the hospital's counter and is subsequently arranged for a consultation with doctor agent Robert Thompson, who is a dermatologist. After undergoing the prescribed medical examination, Mr. Morgan receives a diagnosis and medication. He goes back home to rest and monitor the improvement of his condition. Mr. Morgan needs to go to {\it Agent Hospital} again if he fails to recover after several days.}
\label{fig:closed_circle}
\end{figure*}

\section*{Simulacrum Construction}
Inspired by Smallville \cite{Park:23:Smallville} , we design a hospital sandbox simulation environment using the map editor Tiled \cite{Tiled} and the web game development framework Phaser \cite{Phaser}. As shown in Figure \ref{fig:overview}, there are 16 functional areas in {\it Agent Hospital} such as triage station, registration desk, waiting area, consultation rooms, examination room, pharmacy, and follow-up room.

In {\it Agent Hospital}, we distinguish between two types of autonomous agents: patient agents and medical professional agents. Each agent has distinct demographic information. 
As patient agents may get sick, they have additional information about medical history. For example, Kenneth Morgan is a male patient agent with an age of 55. According to his medical history, he has a hypertension problem. 
Medical professional agents, which include doctors and nurses, have additional information about skills and duties. For example, Robert Thompson is a male dermatologist agent with an age of 46. Proficient in performing skin surgeries, his duty is to diagnose and treat adult patients with a broad range of skin illnesses.
In {\it Agent Hospital}, there are 42 doctor agents and four nurse agents. To simplify the simulation, we assume that medical professional agents will not get sick.

\begin{figure*}[!t]
    \centering
    \includegraphics[width=1.0\textwidth]{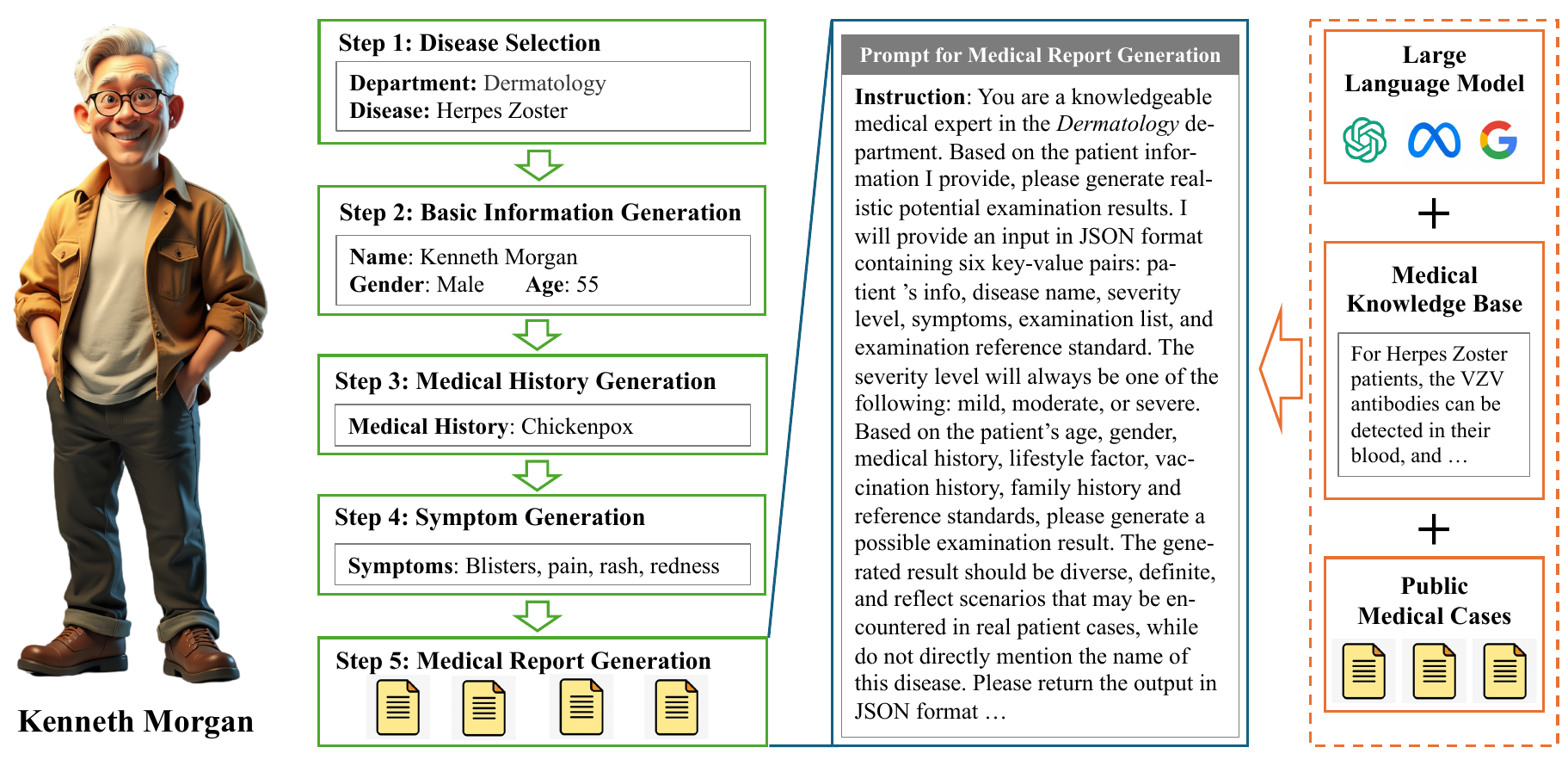}
    \caption{\textbf{Automatic generation of patient agents}. During the closed cycle of treating illness, the simulacrum generates patient agents automatically by coupling large language model with medical knowledge base. After choosing a disease, our method generates the patient's basic information, medical history,  symptoms, and medical examination reports sequentially. Such patient agents are critical for enabling doctor agents to evolve in \textit{Agent Hospital}.}
\label{fig:generation}
\end{figure*}

The simulation process is driven by events in which patient, nurse, and doctor agents get involved. There are eight main types of events in {\it Agent Hospital}:
\begin{enumerate}
\item {\it Disease Onset}. As shown in Figure \ref{fig:closed_circle}, Kenneth Morgan wakes up and finds that his skin becomes red, painful, and blistered. He decides to go to {\it Agent Hospital} to seek medical attention.
\item {\it Triage}. Kenneth Morgan arrives at the triage station and describes his symptoms to the nurse agent Katherine Li, who tells him to register for the dermatology department.
\item {\it Registration}. Kenneth Morgan proceeds to register at the registration counter with the help of the nurse agent Alexander Davis. Then, he goes to the designated area and waits for a consultation.
\item {\it Consultation}. After arriving at the consultation room, Kenneth Morgan describes his symptoms to the dermatologist agent Robert Thompson, who determines the need for a medical examination.
\item {\it Medical Examination}. Kenneth Morgan undergoes the medical test in the examination room. The nurse agent Jessica Chen gives him the report of the examination results.
\item {\it Diagnosis}. Kenneth Morgan goes back to the consultation room with the report. Robert Thompson provides a diagnosis and prescribes the medication after reviewing the examination results.
\item {\it Medicine Dispensary}. Kenneth Morgan goes to the hospital pharmacy, gives the prescription to the nurse agent Andrew Jackson, and picks up his medication.
\item {\it Convalescence}. Kenneth Morgan returns home to commence his recovery. He will provide feedback or updates on his health condition for follow-up actions.
\end{enumerate}
In addition, we design an extra event ``{\it Reading Books}'' for doctor agents: they proactively accumulate knowledge by reading medical books outside of work hours. This is beneficial for integrating medical knowledge and expertise. 

In the real world, it is difficult for human doctors to know whether the treatment plans they prescribed help patients recover or not because many patients do not provide feedback. Fortunately, it is much easier to obtain such feedback and form a closed cycle in {\it Agent Hospital}, making it possible for doctor agents to evolve over a long time. 

\section*{Agent Evolution}
\begin{figure*}[!t]
    \centering
    \includegraphics[width=0.81\textwidth]{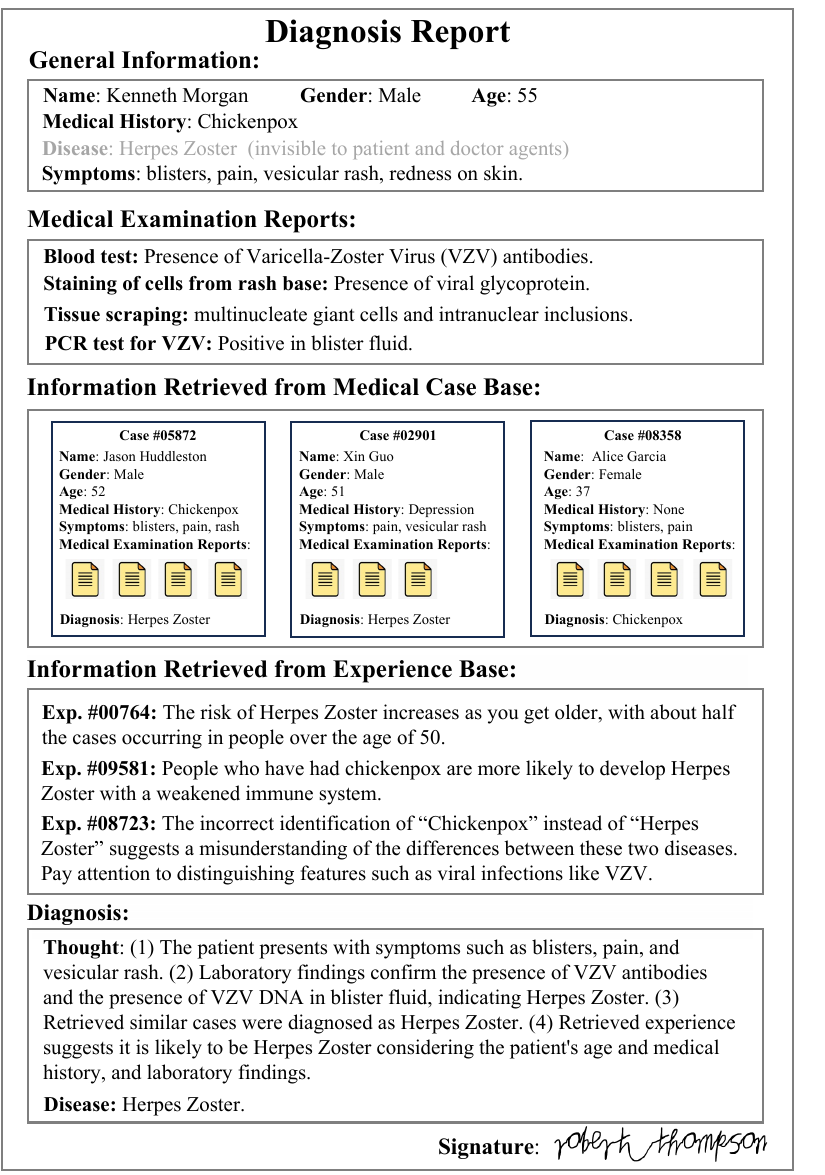}
    \caption{\textbf{An example illustrating how a doctor agent diagnoses a patient agent}. Patient agents, diseases, symptoms, and medical examination results are generated by the simulacrum automatically. Doctor agents diagnose patent agents based on the medical data and update their medical case base or experience base.}
\label{fig:example}
\end{figure*}

In {\it Agent Hospital}, doctor agents evolve mainly by treating patient agents. We refer to our method for agent evolution as {\it MedAgent-Zero}. By ``Zero'', we mean that it does not use any manually labeled data. Instead, it relies only on synthetic medical data generated by the virtual world. {\it MedAgent-Zero} consists of two key steps: {\it patient agent generation} and {\it doctor agent evolution}.  

\subsection*{Patient Agent Generation}
As shown in Figure \ref{fig:generation}, patient agents in {\it Agent Hospital} can be automatically generated by coupling large language models with medical knowledge bases. Given a chosen disease, {\it MedAgent-Zero} first generates the basic information of the patient agent such as name, gender, and age. Learning from the medical knowledge base that people over the age 50 are more likely to contract Herpes Zoster, the large language model could set the age of the patient agent to 55. The model also generates the medical history by adhering to the fact that people who have had chickenpox are more likely to develop Herpes Zoster. Given the disease, basic information, and medical history, the model generates a list of symptoms for the patient agent. Finally, medical examination reports are automatically generated based on medical knowledge on Herpes Zoster. {\it MedAgent-Zero} further uses a quality control agent to ensure that the medical data of a generated patient agent adheres to medical knowledge base.

Patient agent generation is the cornerstone of doctor agent evolution because it can in principle provide an unlimited number of patients for training doctor agents. More importantly, we can easily control the distribution of patient agents in terms of gender, age, country, and disease, making it possible to simulate any patient cohort of interest.

\subsection*{Doctor Agent Evolution}
Given a patient agent, a doctor agent needs to make correct decisions on medical examination, diagnosis, and prescription of medications. Note that only the basic information, medical history, and symptoms of the patient agent are visible to the doctor agent. As our work uses proprietary LLMs as the base model of doctor agents, which are frozen during training and inference, we add two important modules to support agent evolution: {\it medical case base} and {\it experience base}. 

As shown in Figure \ref{fig:example}, after Kenneth Morgan describes his symptoms and submits his medical examination reports, Robert Thompson first retrieves similar cases from the medical case base. For example, the most similar one is case \#05872, in which a patient agent named Jason Huddleston with similar medical history, symptoms, and examination reports was correctly diagnosed as Herpes Zoster. This case can serve as an important reference for Robert Thompson to diagnose the current patient agent. Then, Robert Thompson retrieves rules applicable to the current case from the experience base. For example, the most relevant rule indicates that people over the age of 50 are likely to contract Herpes Zoster. Given the gender, age, medical history, symptoms, medical examination reports, and information retrieved from the medical case base and experience base, Kenneth Morgan is diagnosed with Herpes Zoster and Robert Thompson explains why he makes the decision.

Medical case base and experience base grow with the increase of patient agents being treated. If a doctor agent has successfully treated a patient agent, the case will be added to the medical case base. Otherwise, the doctor agent needs to compare its decisions with the ground-truth decisions and reflect to come up with a rule to avoid making the same mistake again in a way similar to tuning-free rule accumulation \cite{Yang:23:Rule_Accumulation}. If the doctor agent can successfully treat the patient agent by using this rule, the rule will be added to the experience base. Otherwise, the rule will be discarded. 

\begin{figure*}[!t]
    \centering
    \includegraphics[width=0.95\textwidth]{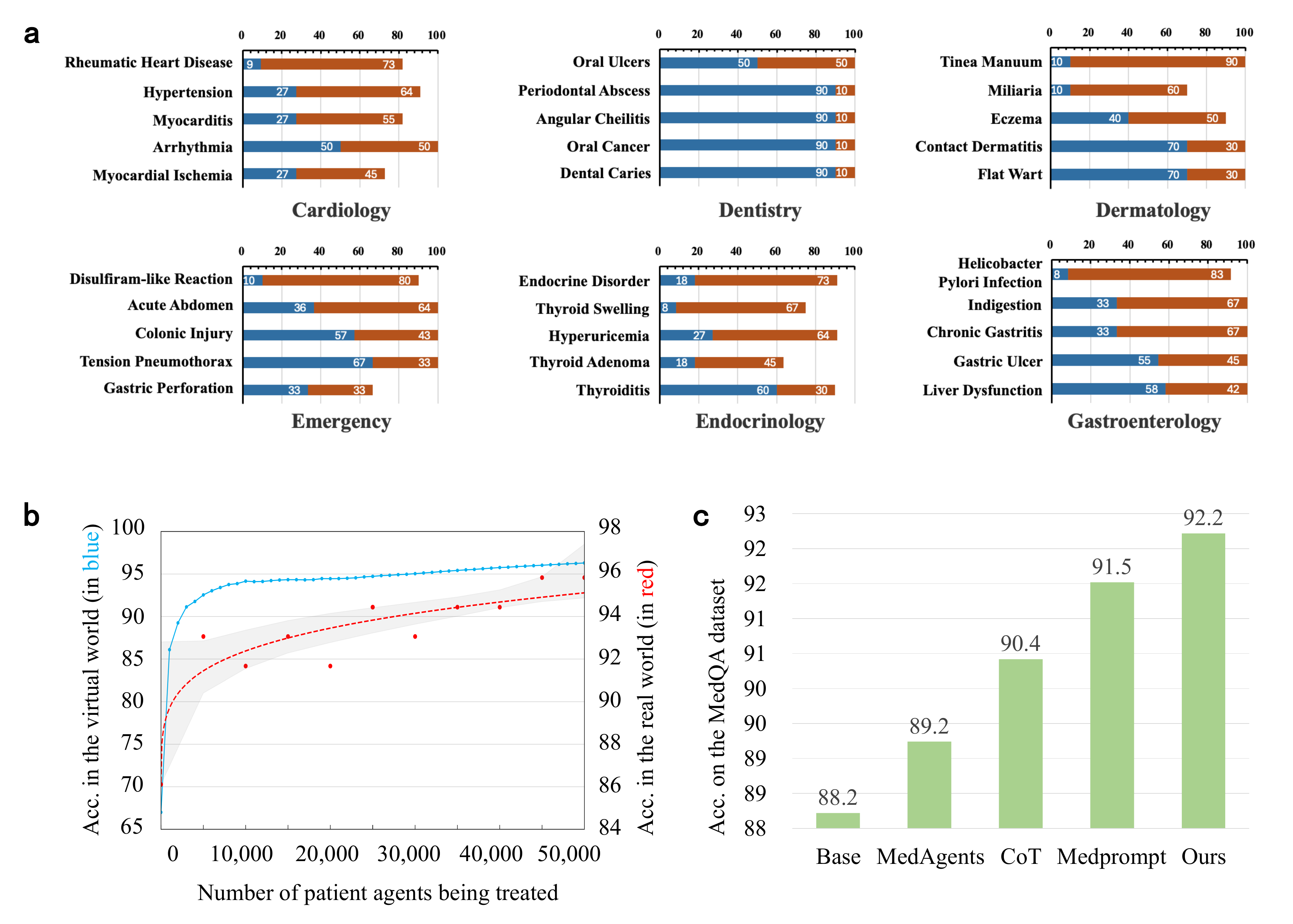}
    \caption{\textbf{Evaluations in the virtual and real worlds}. \textbf{a}, Diagnostic accuracy improvements after doctor evolution over six departments. \textbf{b}, Doctor agents can keep improving  over time both in the virtual and real worlds by treating patient agents without the need to label data manually. In the virtual-world evaluation, we report the accuracy on diagnosing respiratory diseases for patient agents. In the real-world evaluation, we report the accuracy on answering questions related to respiratory diseases in the MedQA dataset. \textbf{c}, Our method outperforms existing methods on the MedQA dataset with GPT-4o as the base model.}
\label{fig:experiments}
\vspace{-5mm}
\end{figure*}

\section*{Scaling Laws in Evolution}

An important question we try to answer is whether doctor agents can keep improving their capabilities with the increase of the number of patient agents being treated. We refer to this question as {\it scaling laws} \cite{Kaplan:20:Scaling_Laws} {\it in evolution}.
In the virtual world of {\it Agent Hospital}, we define three medical tasks to assess the capabilities of doctor agents: {\it medical examination selection}, {\it diagnosis}, and {\it treatment plan recommendation}. In the medical examination selection task, the doctor agent selects appropriate medical tests that the patient agent needs to undergo given symptoms. In the diagnosis task, the doctor agent provides a diagnosis after reviewing symptoms and examination results. In the treatment plan recommendation task, the doctor agent decides on an appropriate treatment plan for the patient agent.
There are 32 medical departments in {\it Agent Hospital} , covering more than 300 diseases. For each clinical department, we construct a training set and a test set for each task, which contain 20,000 and 200 patient agents respectively.

As shown in Figure \ref{fig:experiments}a, doctor agent evolution significantly improves diagnostic accuracy for six departments.  For example, in the cardiology department, the diagnostic accuracy of rheumatic heart disease is only 9\% in the beginning when only the base model GPT-3.5 is used. After doctor agent evolution, the accuracy dramatically increases to 82\%. Similar results have been observed for other departments, suggesting our method {\it MedAgent-Zero} might be generalizable to all diseases.

What will happen if doctor agents treat more patient agents? The blue curve in Figure \ref{fig:experiments}b shows the diagnostic performance of the doctor agent in the respiratory department after treating 50,000 patient agents. In the beginning, the diagnostic accuracy is around 66\%. The accuracy increases dramatically when the doctor agent is treating the first 10,000 patient agents as the medical case base and experience base quickly grow. The improvement slows down but still increases steadily with more patient agents coming. We observe similar curves for other departments. To save training costs, we use at most 20,000 patient agents for each department to train doctor agents in the following experiments.

Figure\ref{fig:experiments} c shows the results for the entire MedQA dataset. All methods use GPT-4o as the base model. We find that our method outperforms the state-of-the-art methods on medical agents such as MedAgents \cite{Tang:23:MedAgents}, CoT \cite{Wei:22:CoT}, and Medprompt \cite{Nori:23:Medprompt}. This finding is encouraging since we do not use the training data of MedQA.

\section*{Alignment between Virtual and Real Worlds}

Another important question is whether the expertise doctor agents acquired in the virtual world can be applicable to the real world. We refer to this problem as {\it the alignment between the virtual world and the real world}. This problem is very important because {\it Agent Hospital} is both a medical world simulator and a time accelerator, which can enable doctor agents to quickly evolve in the virtual world and provide high-quality medical services in the real world. If the medical skills learned from the virtual world can be used to solve real-world medical problems, it is possible to create superhuman AI doctors just like what AlphaGo Zero did in playing Go because the virtual world can be easily customized to accommodate various real-world scenarios and generate an unlimited amount of medical data for training AI doctors.

Figure \ref{fig:experiments}b reports preliminary positive results on the alignment problem. The red dots denote the accuracies of doctor agents answering questions related to respiratory diseases in the MedQA dataset at different sampled times. The red dotted curve is the trend curve, which suggests that the accuracy in the real world generally increases with the increase of  diagnostic accuracy in the virtual world.

\section*{Related Work}
\subsection*{LLM-powered Medical Agents}
As LLMs have demonstrated significant intelligence in reasoning and instruction-following, abundant efforts have been made in training foundation models and vertical-domain models. Recent studies show that LLM-powered agents are as strong as humans to some extent, as they can act in an environment, have their own memory, and know how to make use of external tools~\cite{huang2024understanding}. These advancements have been further extended to LLM-based multi-agent systems, which enhance reasoning and planning by simulating human activities and optimizing the collective power of multiple agents \cite{li2024survey}. 

In the medical domain, various research efforts in recent years have focused on building foundational medical models such as Google’s MedPalm series~\cite{Tu:24:MedPaLM_Med,  singhal2023towards}. Since 2023, there has also been a growing interest in developing medical agents. For instance, the MedAgents framework employs a multidisciplinary collaborative approach to enhance the performance of LLMs in zero-shot medical reasoning \cite{Tang:23:MedAgents}, which highlights the versatility and utility of LLM agents in handling specialized terminology and complex reasoning in medical applications \cite{qiu2024llm}. Recent studies are also exploring ways to enable medical agents to accomplish more complex tasks through collaboration~\cite{Kim:24:MDAgents, Li:24:Mmedagent}. However, existing work lacks the integration of personalized memory for agents, limiting their capabilities to simple role-playing techniques to activate LLM functionalities. Consequently, these agents struggle to accumulate unique experience and evolve continuously during interactions, unlike human doctors. Compared with these studies, our study provides a new paradigm to train powerful medical agents.

\subsection*{Self-Improving LLM-powered Agents}
LLMs have achieved multiple breakthroughs through methodologies such as pre-training~\cite{devlin-etal-2019-bert}, fine-tuning~\cite{raffel2020exploring}, and other forms of human-supervised training~\cite{ouyang2022training}. However, current LLMs and agents may encounter limitations in performance as task complexity and diversity escalate. Existing training paradigms, which require the use of extensive data corpora or heavy human supervision, are deemed costly. Therefore, the development of self-evolutionary approaches has gained momentum. These approaches enable LLM-powered agents to autonomously acquire, refine, and learn through self-evolving strategies.

LLM-powered agents can engage in a reflection process during solution generation to facilitate self-improving. LSX~\cite{stammer2023learning} introduces two interconnected modules working in tandem to evolve: a learner module that executes a foundational task and a critic module that evaluates the quality of explanations provided by the learner. Furthermore, SelfEvolve~\cite{jiang2023selfevolve} and LDB~\cite{ zhong2024ldb} enhance an agent’s capability in code generation by enabling it to reflect on and learn from feedback generated during operation. Through such reflective processes, agents can self-evolve, refine their methodologies, and thus achieve improved performance. However, current studies on agent evolution predominantly concentrate on isolated tasks, with insufficient integration into environments, which are vital for humans to evolve their capabilities. Some embodied AI studies propose to enable LLM-powered agents to evolve in the real world~\cite{bovo2024embardiment}, but SEAL proposes to construct a virtual environment to accelerate the evolution of medical agents.

\subsection*{Simulacrum Construction for LLM-powered Agents}
Recent research initiatives have leveraged LLMs to replicate real-world dynamics. In multiple fields such as epidemiology, sociology, and economics, researchers are utilizing LLM-powered agents to simulate human decision-making, leading to many exciting emergence phenomena
in various domains~\cite{mou2024individual}.

Smallville~\cite{Park:23:Smallville} is one of the earliest works to explore the use of agents in social simulation. It creates a virtual town to simulate human daily life and observed that the agents exhibited behaviors resembling those of humans. 
Li et al.~\cite{li2024econagent}  construct a macroeconomic system to simulate individual work and consumption behaviors, which is more powerful than previous simulation strategies in showing economic phenomena.
However, existing studies primarily focus on simulating open societies to verify or explore new social behavior theories~\cite{chawla2023selfish, li2023camel}, or on replicating specific workflows within predefined scenarios to enhance agent performance~\cite{qian2024chatdev, chen2023agentverse}. In contrast, our study proposes a novel approach: constructing a simulacrum of a closed-cycle scenario and enabling agents to evolve through interactions within the virtual environment.

\section*{Discussion}
We have presented a simulacrum of hospital called {\it Agent Hospital} for generating AI patients and training AI doctors, both of which might have profound impacts on medical AI. 
On the one hand, AI patients can be applied to a variety of scenarios such as modeling disease onset and progression, simulating patient cohorts for specific diseases and regions, training medical students and junior doctors, representing human patients to consult multiple doctors concurrently, and preserving privacy for human patients.
On the other hand, AI doctors have the potential to revolutionize the healthcare industry since they can learn how to treat diseases quickly  by utilizing vast amounts of data beyond human doctors can perceive and handle, help to reduce workload and improve efficiency for human doctors, and create a more equitable and effective healthcare system.

Different from large language models that are trained on manually labeled data without any environment, the SEAL paradigm we propose combines the merits of AlphaGo Zero \cite{Silver:17:AlphaGo-Zero} and Smallville \cite{Park:23:Smallville}: building a simulacrum of the real-world scenario of interest and enabling autonomous agents to evolve in the simulacrum without using manually labeled data. An important characteristic of SEAL is to use large language models coupled with domain knowledge bases to generate training data. We believe that this paradigm can be adopted in many other applications that involve multiple agents such as virtual court \cite{Chen:24:AgentCourt}. In the future, we will further improve our method to overcome the following limitations: the base model is frozen and non-evolvable, AI doctors can only recommend high-level treatment plans, and lack of consultation of doctors from different departments.

Despite the benefits resulted from {\it Agent Hospital}, we must carefully deal with ethical considerations and societal impact. AI doctors are prone to inherit and amply biases present in the training data, leading to discriminatory outcomes. To address this problem, we will develop debiasing techniques to control the distribution of generated AI patients to ensure fairness and equity in AI-driven healthcare. Furthermore, AI doctors are required to provide detailed chains of thoughts to ensure transparency and accountability, build trust, and mitigate potential harms to human patients. The development and application of {\it Agent Hospital} will strictly comply with current laws, regulations, and ethical constraints. Our long-standing goal is to use AI technology to provide cheap, accessible, and high-quality medical services to the public.

\section*{Acknowledgments} This work is supported by the National Natural Science Foundation of China (No. 61925601, 62372260, 62276152) and AI Industry Research Innovation Center (AIRIC), Wuxi Research Institute for Applied Technologies, Tsinghua University. 

\section*{Author Contributions} 
Yang Liu led the project, coined the term ``Agent Hospital'' and designed the overall research and development framework. Weizhi Ma organized the research and managed the overall project progress, including collecting datasets, designing models and algorithms, training doctor agents, implementing the online system, and writing the manuscript, among others. Junkai Li, Weitao Li, Weizhi Ma, and Yang Liu proposed the MedAgent-Zero method, which was then implemented by Junkai Li and Weitao Li. Yunghwei Lai, Weizhi Ma, and Yang Liu proposed the generation algorithm of patient agents, and Yunghwei Lai was responsible for the implementation. The experiments were conducted by Junkai Li, Jingyi Ren, Weitao Li, Yunghwei Lai, and Xinhui Kang. Meng Zhang and Siyu Wang contributed to the construction of the simulacrum system and the collection of medical knowledge bases. Peng Li contributed to the idea of evolvable agents and participated in the discussion. Ya-Qin Zhang advised the project, participated in the discussion, and offered insightful suggestions for the development of Agent Hospital. Weizhi Ma and Yang Liu proofread the whole manuscript.

\bibliography{references}
\renewcommand\refnamemethods{} 

\newpage
\section*{Appendix}

\section*{A. Details of Medical Datasets and Knowledge Bases}

\subsection*{A.1 Categorization of Medical Departments}
In \textit{Agent Hospital}, we aim to cover all hospital departments' capabilities by training corresponding doctor agents, as various human doctors in a real hospital. First, we need to determine the department categories. Due to differences in department setups across various hospitals, it is challenging to find a standardized approach. Therefore, we adopted an LLM-powered selection method by using GPT-4 to classify all test questions in the MedQA dataset by departments, which is a prompt-driven approach to generate a full list of departments that can broadly cover all capabilities that are necessary for physicians.

Then, we get 32 medical departments, which can be classified into two types: clinical departments and non-clinical departments. The reason we distinguish between these two types of departments is that doctors in clinical departments will interact with patients directly, but non-clinical departments will not. 
Finally, there are a total of 21 clinical medical departments for patient treatment, such as the respiratory department, emergency department, and so on. The complete list of clinical medical departments is shown in Table \ref{table:clinical_departments}. 
Besides, there are also 11 non-clinical medical departments, which focus on the foundational aspects of medicine rather than direct patient care/treatment. The full list of these non-clinical departments is summarized in Table \ref{table:non_clinical_departments}. 

\begin{table}[H]
\caption{21 clinical medical departments in \textit{Agent Hospital}.} \label{table:clinical_departments}

\resizebox{\columnwidth}{!}{
\begin{tabular}{|c|c|c|ccc}
\hline
\textbf{\begin{tabular}[c]{@{}c@{}}Cardiology\\ Department\end{tabular}}                    & \textbf{\begin{tabular}[c]{@{}c@{}}Dentistry\\ Department\end{tabular}}    & \textbf{\begin{tabular}[c]{@{}c@{}}Dermatology\\ Department\end{tabular}}    & \multicolumn{1}{c|}{\textbf{\begin{tabular}[c]{@{}c@{}}Emergency\\ Department\end{tabular}}}    & \multicolumn{1}{c|}{\textbf{\begin{tabular}[c]{@{}c@{}}Endocrinology \\ Department\end{tabular}}}                           & \multicolumn{1}{c|}{\textbf{\begin{tabular}[c]{@{}c@{}}Gastroenterology \\ Department\end{tabular}}} \\ \hline
\textbf{\begin{tabular}[c]{@{}c@{}}General Surgery\\ Department\end{tabular}}               & \textbf{\begin{tabular}[c]{@{}c@{}}Hematology \\ Department\end{tabular}}  & \textbf{\begin{tabular}[c]{@{}c@{}}Immunology \\ Department\end{tabular}}    & \multicolumn{1}{c|}{\textbf{\begin{tabular}[c]{@{}c@{}}Infectious \\ Department\end{tabular}}}  & \multicolumn{1}{c|}{\textbf{\begin{tabular}[c]{@{}c@{}}Nephology \\ Department\end{tabular}}}                              & \multicolumn{1}{c|}{\textbf{\begin{tabular}[c]{@{}c@{}}Neurology \\ Department\end{tabular}}}        \\ \hline
\textbf{\begin{tabular}[c]{@{}c@{}}Obstetrics and \\ Gynecology \\ Department\end{tabular}} & \textbf{\begin{tabular}[c]{@{}c@{}}Oncology \\ Department\end{tabular}}    & \textbf{\begin{tabular}[c]{@{}c@{}}Ophthalmology \\ Department\end{tabular}} & \multicolumn{1}{c|}{\textbf{\begin{tabular}[c]{@{}c@{}}Orthopedics \\ Department\end{tabular}}} & \multicolumn{1}{c|}{\textbf{\begin{tabular}[c]{@{}c@{}}Otolaryngology \\ Department\end{tabular}}} & \multicolumn{1}{c|}{\textbf{\begin{tabular}[c]{@{}c@{}}Pediatrics \\ Department\end{tabular}}}       \\ \hline
\textbf{\begin{tabular}[c]{@{}c@{}}Psychiatry \\ Department\end{tabular}}                   & \textbf{\begin{tabular}[c]{@{}c@{}} Respiratory\\ Department\end{tabular}} & \textbf{\begin{tabular}[c]{@{}c@{}}Urology \\ Department\end{tabular}}       & \textbf{}                                                                                       & \textbf{}                                                                                                                   & \textbf{}                                                                                            \\ \cline{1-3}
\end{tabular}
}
\end{table}

\begin{table}[H]
\caption{11 non-clinical medical departments in \textit{Agent Hospital}.} \label{table:non_clinical_departments}
\resizebox{\columnwidth}{!}{
\begin{tabular}{|c|c|c|c|ccc}
\hline
\textbf{Anatomy}   & \textbf{Anesthesiology} & \textbf{Biochemistry} & \textbf{Genetics}                                                       & \textbf{\begin{tabular}[c|]{@{}c@{}}Internal \\ Medicine\end{tabular}} & \multicolumn{1}{|c|}{\textbf{Microbiology}} & \multicolumn{1}{c|}{\textbf{Pathology}} \\ \hline
 \textbf{Pharmacology}   & \textbf{Physiology}   & \textbf{\begin{tabular}[c]{@{}c@{}}Preventive \\ Medicine\end{tabular}} & \textbf{Radiology}                                         & \textbf{}          & \textbf{}   & \textbf{}                                  \\ \cline{1-4}
\end{tabular}
}
\end{table}

Note that as non-clinical departments do not interact with AI patients, the only way to enhance capabilities is through learning without practice for the non-clinical departments, leading to slight differences in doctor agent training. 

\subsection*{A.2 Disease Knowledge Collection for Clinical Departments}

Disease knowledge is necessary for the generation of accurate, diverse medical records in various clinical departments, which is vital for the construction of patient agents. We chose to use disease information from the authoritative Baidu Health Encyclopedia\footnote{https://jiankang.baidu.com/widescreen/entitylist} for generating medical records. On this website, we can easily obtain comprehensive information about each disease, including causes, symptoms, and possible treatment options. This allows us to quickly organize knowledge about various diseases for simulation. In Figure \ref{fig:COVID-19 data information}, we present some disease information related to COVID-19, note that to avoid redundant information, we only used relevant information such as symptoms, clinical manifestations in medical examinations, and treatment plans, but not all disease knowledge.

\FloatBarrier
\begin{figure}[H]
\begin{mdframed}[innerleftmargin=10pt,innerrightmargin=10pt]
\underline{\textbf{COVID-19}} \vspace{3mm} \\
\textbf{Symptoms:} dry throat, sore throat, fever, smell taste loss, runny nose, the central nervous system involvement, difficulty in breathing, hypoxemia, acute respiratory distress syndrome, sepsis shock, refractory metabolic acidosis, coagulopathy, and multiple organ failure.\vspace{3mm}\\
\textbf{Examination Results:} 
\begin{itemize}[leftmargin=2em,    
                itemsep=0.5em,    
                topsep=0.5em,     
                parsep=0pt,       
                labelsep=0.5em]   
      \item \textbf{Blood Test}: In the early stage of the disease, the total number of peripheral blood white blood cells was normal or decreased, and the lymphocyte count was decreased. Some patients may have increased liver enzymes, lactate dehydrogenase, muscle enzymes, myoglobin, troponin, and ferritin. In most patients, C-reactive protein (CRP) and erythrocyte sedimentation rate were increased, and procalcitonin was normal. In severe and critical patients, D-dimer was increased, peripheral blood lymphocytes were progressively decreased, and inflammatory factors were increased.
      \item \textbf{Chest X-ray Exam}: Chest imaging examination showed multiple small patchy shadows and interstitial changes in the early stage, which were obvious in the outer lung zone. Then, it develops into multiple ground-glass opacities and infiltrations in both lungs. In severe cases, lung consolidation may occur, and pleural effusion is rare. In MIS-C, enlarged heart shadow and pulmonary edema are seen in patients with cardiac dysfunction.
\end{itemize}
\vspace{3mm}
\textbf{Treatment Plan:} 
\begin{itemize}[leftmargin=2em,    
                itemsep=0.5em,    
                topsep=0.5em,     
                parsep=0pt,       
                labelsep=0.5em]   
    \item \textbf{Mild}: Rest in bed, strengthen supportive treatment, ensure adequate energy and protein intake, supplement vitamins, trace elements, and other nutrients; Timely administration of ritonavir tablets or ambavir and romisivir injection.
    \item \textbf{Moderate}: Timely physical cooling, drug antipyretic, prone position treatment, timely delivery of azvudine, monolavir capsule drug treatment.
    \item \textbf{Severe}: Treatment was given in the standard prone position for no less than 12 hours per day. Respiratory support, circulatory support, and timely administration of intravenous human immunoglobulin for COVID-19.
\end{itemize}
\end{mdframed}
\caption{Part of preprocessed medical knowledge about the COVID-19.}
\label{fig:COVID-19 data information}
\end{figure}
\FloatBarrier

Based on our pilot experiments, hundreds of generated patients for each disease are required for the training of doctor agents. However, generating a large number of virtual patients covering all the diseases mentioned above requires a significant amount of time and resources. Besides, from a medical standpoint, the diagnosis and knowledge acquisition of common diseases are more important. Thus, we referred to a list of common diseases from websites such as DXY \footnote{https://dxy.com/diseases/6948}, which is an authoritative online medical website in China. This allows us to identify the most typical diseases for each clinical department for data generation and subsequent computational processes. Based on the common disease list here, we supplement the departments with fewer diseases using the common disease information provided by GPT-4. Finally, there are 339 diseases across the 21 clinical departments adopted in \textit{Agent Hospital}.

\section*{B. Implementation Details}
The generation of synthetic medical cases and patient agents plays a critical role in our work. Here, we provide a detailed introduction focusing on the self-evolution details of doctor agents, which includes three main aspects: doctor agent response generation, medical case base construction (i.e., learning from success), and experience base accumulation (i.e., learning from failures). 

\subsection*{B.1 Doctor Agent Response Generation}
As LLMs inherently possess strong language capabilities, we focus on enhancing the critical medical decision-making abilities of doctor agents such as determining examination options and providing diagnoses. For any of the above tasks, we design a basic question prompt structure to guide the doctor agents' judgments, which includes: 1) \textbf{Instruction}: Clearly define the current doctor agent’s identity and role. 2) \textbf{Patient Information}: This varies depending on the specific task. For example, examination results are not provided when selecting examination items, but they are included when making a diagnosis. 3)~\textbf{Candidate Choices}: Since primary examinations and diseases for each department can be enumerated, we construct a candidate list. If none of the candidates is suitable, doctor agents leverage their generative capabilities to propose new options. 4) \textbf{Personal Experience}: Each doctor agent has its own medical case base and experience base, which are utilized during reasoning through Retrieval-Augmented Generation (RAG)~\cite{}. Based on such a structured prompt input, our medical agent can make medical decisions with detailed reasoning steps.

An RAG module is adopted to select the most helpful information from medical case base and experience base for the current question. To be more specific, during the storage process of medical cases or experience, the corresponding question is also saved for the calculation of relevance with the current question. When using RAG, the current question is vectorized using the same encoder as previous questions. We use cosine similarity as a metric to find the top-$n$ related cases and top-$k$ related principles, which are then used in the inference prompt. We choose the text-embedding-ada-002\footnote{https://platform.openai.com/docs/guides/embeddings/embedding-models} model provided by OpenAI as the text encoder, which allows each stored question to be represented as a vector, creating vector databases of medical case base and experience base for RAG, respectively.  

\subsection*{B.2 Medical Case Base Accumulation}
It is beneficial for human doctors to use similar medical cases to help deal with a new case. As a result, we construct a medical case base for each doctor agent to store the successful decisions in a similar way. The medical case base is structured in the format of question-answer pairs, where the question details the medical condition requiring decision-making, and the answer contains the validated response. 

A medical case base can be built in two ways: 
\textbf{1) Patient-Doctor Agent Interaction}. For each generated answer from the doctor agent, the question-answer pair will be added to the medical case base if its answer is correct. As there are various medical tasks, we prefer that each task should have its private medical case base to avoid irrelevant case utilization. The question part of each task is distinct. For example, we record symptoms for the examination task and symptoms as well as examination results for the diagnosis task, respectively.
\textbf{2) Medical Knowledge Learning}. Apart from improving their skills through clinical practice, doctor agents also proactively accumulate knowledge by reading medical documents outside of work hours. To avoid parametric knowledge learning for agents, we propose to reorganize the medical documents into multi-choice questions with the help of LLMs so that they will follow the same format as patient questions to be added to the medical case base.

\subsection*{B.3 Experience Base Reflection, Validation and Refinement}

To enable doctor agents to learn from failures, we propose methods for experience reflection, validation, and refinement. 

The three key components are: \textbf{1) Experience Reflection}. Reflection is vital for doctor agents to come up with experience to avoid making the same mistake again. If the answer is wrong, the doctor agent will compare the wrong answer with the ground-truth answer and come up with a principle. Note that as such principles are in natural language, it is easy to understand, modify, and integrate with human doctors' experience.
\textbf{2)~Experience Validation}. Each principle stored in the experience base will be tested against Q\&A pairs drawn from exemplar cases from medical documents. When a principle is applied to new queries, the diagnostic outcomes are evaluated. If the diagnosis aligns with the expected results, the principle is validated and will be included in the refined experience base. However, if the principle leads to an incorrect diagnosis, it will be discarded. This allows the medical professional agent to apply accumulated knowledge across diverse cases, identifying inconsistencies or areas of improvement in its diagnostic reasoning.
\textbf{3)~Experience Refinement}. As there are different formats of experience that result in retrieval bias, we reformat all principles in experience base with manually selected examples. Note these reformatted principles are also refined by the process above.
To eliminate the influence of noise and maximize the utilization of the experience base, we incorporate additional judgment when utilizing experience. This judgment involves evaluating whether the top-$K$ experience retrieved based on semantic similarity is helpful for the treatment process. 

\section*{C. Supplementary Experiments and Analyses}

\subsection*{C.1 Evolution Performances in the Rest Clinical Departments}

\begin{figure}[!h]
    \centering
    \includegraphics[width=\linewidth]{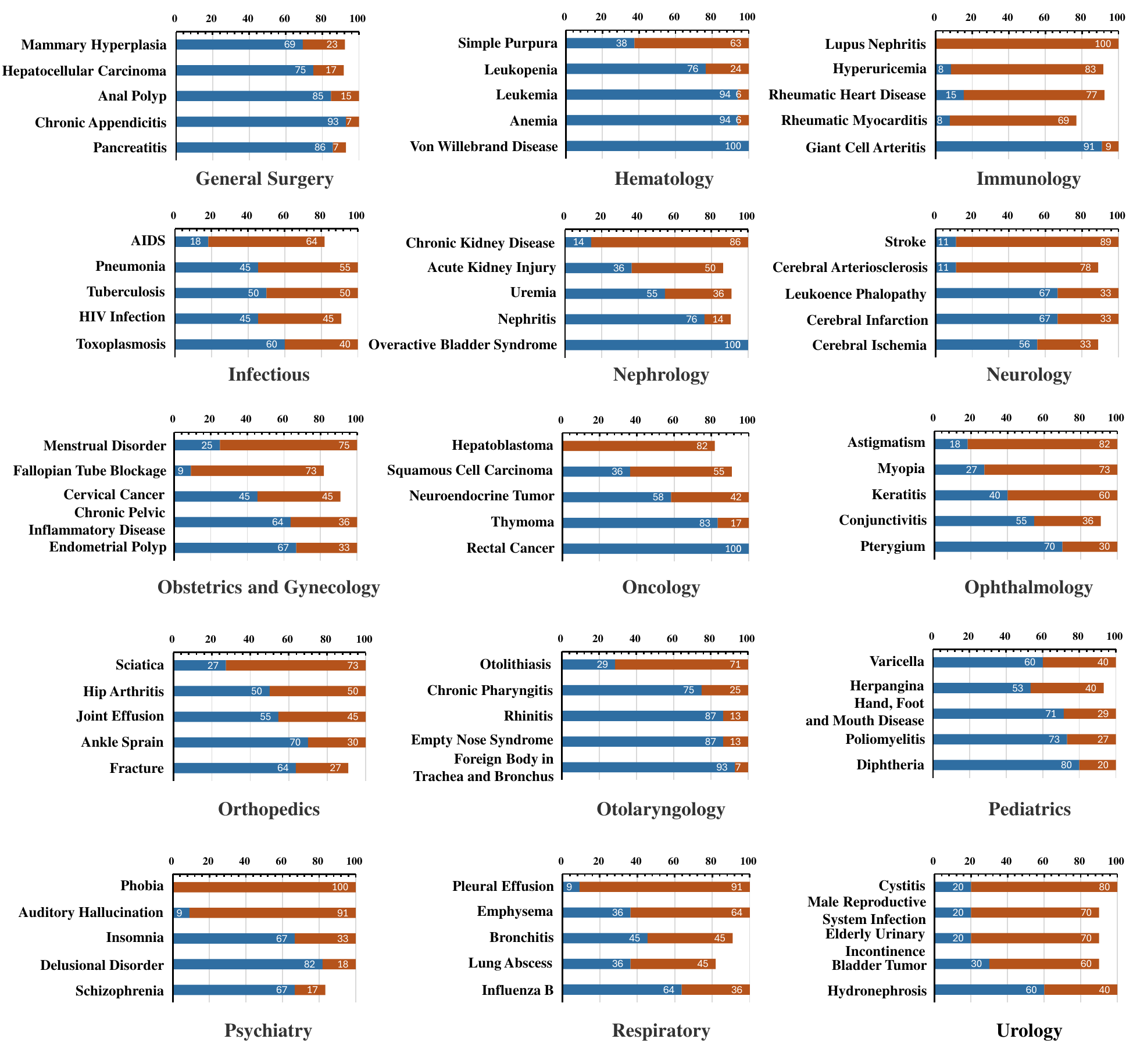}
    \caption{Diagnostic accuracy improvements after doctor evolution on rest clinical departments.}
    \label{fig:case_comparison1}
\end{figure}

We first show the diagnosis accuracy on the rest of the clinical departments in Figure~\ref{fig:case_comparison1}. The diagnosis accuracy improvements across multiple medical departments before and after the integration of the evolved doctor agent based on the proposed \textit{MedAgent-Zero}. 

The figure illustrates consistent accuracy increments for the top five diseases with the highest performance gains across a wide range of conditions, spanning departments such as General Surgery, Infectious Diseases, Hematology, Immunology, Neurology, and Oncology. Each subfigure highlights a notable increase in diagnostic precision, with significant improvements observed in conditions such as Mammary Hyperplasia (General Surgery), Lupus Nephritis (Immunology), Acute Kidney Injury (Nephrology), Cervical Cancer (Obstetrics and Gynecology), and Influenza B (Respiratory). These consistent advancements across diverse domains underscore {\it MedAgent-Zero}'s capability to enhance diagnostic accuracy, particularly for complex or nuanced conditions, highlighting its potential to support and augment medical decision-making.

\begin{figure}[htbp]
    \centering
    \includegraphics[width=0.6\linewidth]{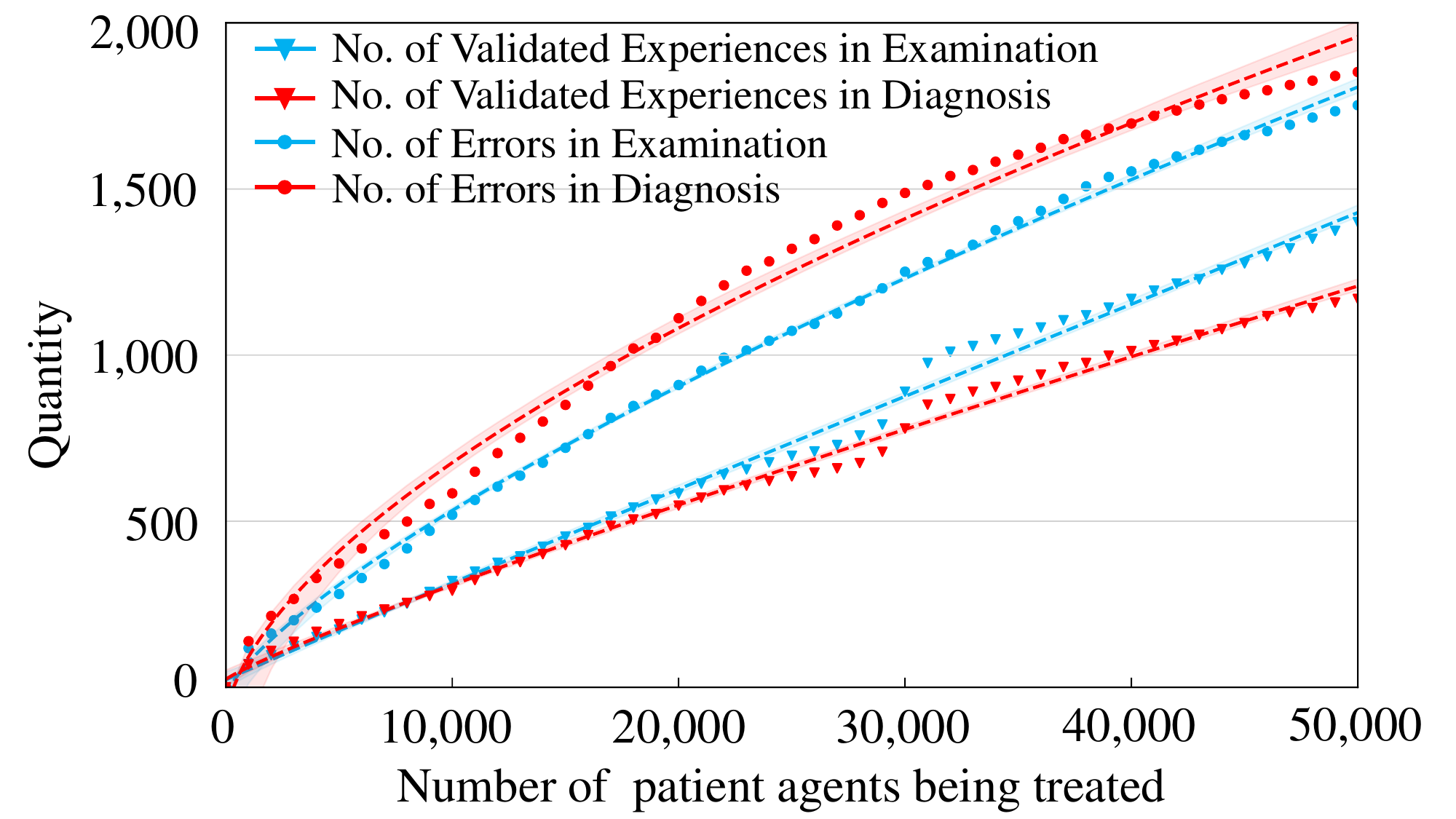}
    \caption{Accumulation of validated experiences during the evolution of doctor agents in examination and diagnosis tasks. An error response refers to an incorrect answer provided by the doctor agents for a given task. A validated experience represents it is reflected and validated. Note that not all errors lead to the reflection of valuable experiences.}
    \label{fig:number of wrong and validated}
\end{figure}

\subsection*{C.2 Analysis of Experience Accumulation}

As introduced in the Method section of the main text, doctor agents will reflect experience from error answers and valid it by themselves. Here we show how the experience accumulated with the increase of interacted patient agents in Figure~\ref {fig:number of wrong and validated}.

As can be seen in the figure, it depicts the accumulation of validated experiences and error responses in the respiratory department during treating 50,000 patient agents, where both the examination and diagnosis tasks are reported. As the number of training samples increases, both validated experiences and error responses gradually rise. As experiences are reflected when doctor agents generate wrong answers and have to pass the validation, the curve of experiences is always below the error curve.
Furthermore, experience accumulation is more efficient in the examination task, evidenced by the consistently higher quantity of validated experiences and the lower number of error responses compared to the diagnosis task. 
This difference may be due to the greater complexity involved in reflecting experiences from the diagnosis task compared to the examination task. 
Besides, experimental results in other department also show similar trends.
Finally, note that the accumulation becomes slower with the increase in patients, indicating that reflecting on new and valid experiences is more difficult than in the beginning. This result is similar to human learning, the more the harder. A slight trend change can be seen around 30,000 patients, after checking the data it may caused by the API update from OpenAI.

\subsection*{C.3 Trends in Precision with Increasing Number of AI patients}

To further validate the changes brought about by increasing the number of AI patients during the training process, we show two other departments, the Cardiology Department and the Nephrology Department, rather than the respiratory Department to analyze the details of the experience accumulation process. We will focus not only on the cumulative accuracy for both the examination and diagnosis tasks but also on the accuracy at each segment (per 1,000 AI patients). In terms of data scale, each department utilized up to 20,000 AI patients.

\begin{figure}[!h]
    \centering
    \includegraphics[width=\linewidth]{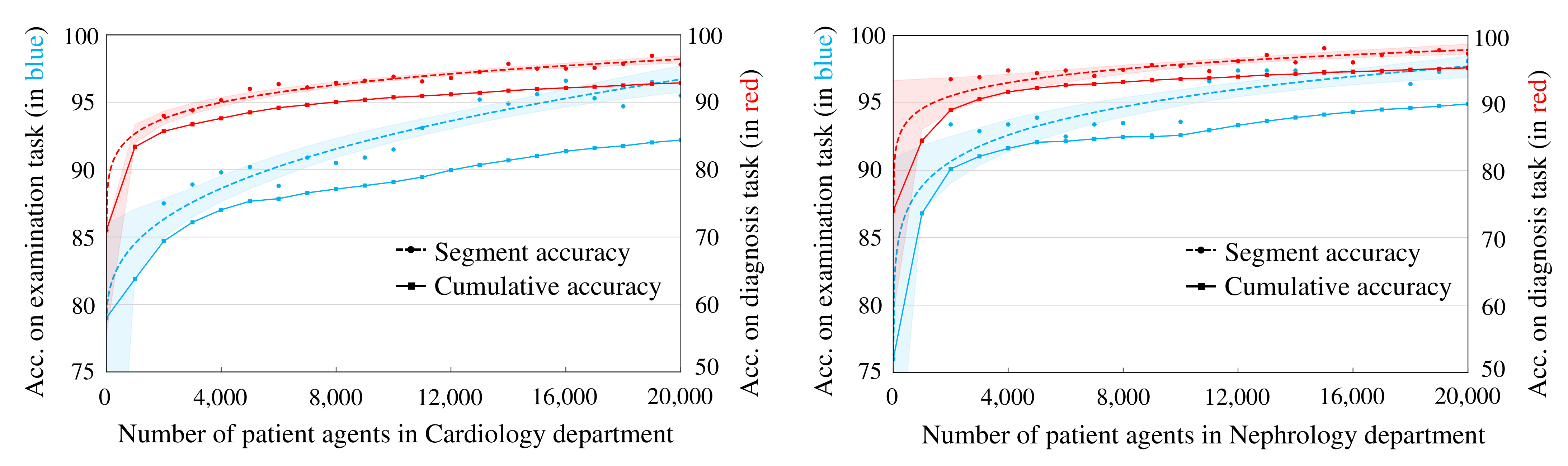}
    \caption{The cumulative and segment accuracy of examination and diagnosis tasks in the Cardiology and Nephrology departments in relation to the increasing number of treated patients. Cumulative accuracy refers to the success rate across all patients treated, whereas segment accuracy represents the success rate calculated for the most recent cohort of 1,000 patients. The dotted lines indicate the fitted curve of segment accuracy, and the shaded areas represent its confidence interval.}
    \label{fig:cumulative and segment accuracy}
\end{figure}

Figure~\ref{fig:cumulative and segment accuracy} illustrates the evolution process of doctor agents in the two departments, which demonstrates that as the number of treated patients increases, the accuracy of both examination and diagnosis tasks improves progressively in both departments, validating that the proposed framework successfully develops evolvable doctor agents in a virtual world with good generalization ability. Furthermore, the cumulative and segment accuracies for both tasks in both departments achieve scores exceeding 90\%, representing an improvement of nearly 25\% compared to the initial state. Notably, these curves show a rapid increase during the treatment of the first 2,000 patients, with the segment accuracy curves maintaining an upward trend throughout the entire process. However, treating more patients is not always better, as we find that there is a marked drop between 12,000 cases and 14,000 cases in the Cardiology department. The reason may be that some unhelpful experience is concluded. However, with more and more experience of high quality, the overall trend is getting better and better. As the precision trends of other departments are similar, so we do not show all the figures here.

\subsection*{C.4 Performance of Evolved Agents Across Clinical Departments with AI Patients}

To verify whether the proposed \textit{MedAgent-Zero} method can achieve consistent improvements across AI patients in different clinical departments, we constructed 20,000 virtual patients for each of the 21 clinical departments mentioned in Section A.1. We then compared the accuracy of the agents before and after evolution. Note that the accuracy here was tested on a separate set of 200 AI patients per department, without overlap with the training data.

\begin{table}[htpb]
\centering
\caption{The results of examination and diagnosis tasks across all clinical departments before and after doctor agent self-evolution. \textit{Original} and \textit{Evolved} refer to the initial and the evolved agents, respectively.}
\label{tab: all department acc before and after evolution internal}
\begin{tabular}{lrrrr}
\hline
\multirow{2}{*}{\textbf{Department}} & \multicolumn{2}{r}{\textbf{Accuracy in Examination}} & \multicolumn{2}{r}{\textbf{Accuracy in Diagnosis}} \\ \cline{2-5} 
& \textbf{Original} & \textbf{Evolved} & \textbf{Original} & \textbf{Evolved}      \\ \hline
Cardiology       & 52.50\% & 96.00\%  & 68.00\% & 93.50\%  \\
Dentistry        & 49.00\% & 99.00\%  & 89.50\% & 98.00\%  \\
Dermatology      & 55.50\% & 94.50\%  & 77.00\% & 93.00\%  \\
Emergency        & 60.50\% & 99.00\%  & 71.50\% & 94.00\%  \\
Endocrinology    & 85.50\% & 99.00\%  & 69.00\% & 93.00\%  \\
Gastroenterology & 39.00\% & 98.00\%  & 69.00\% & 98.50\%  \\
General Surgery          & 84.50\% & 100.00\% & 87.00\% & 97.00\%  \\
Hematology       & 88.50\% & 99.50\%  & 92.00\% & 100.00\% \\
Immunology     & 71.50\% & 100.00\% & 67.50\% & 87.50\%  \\
Infectious       & 71.00\% & 99.50\%  & 76.00\% & 96.50\%  \\
Nephrology       & 62.50\% & 97.50\%  & 76.50\% & 96.50\%  \\
Neurology        & 84.50\% & 99.00\%  & 74.50\% & 88.00\%  \\
Obstetrics and Gynecology  & 88.50\%  & 100.00\% & 78.50\% & 95.00\% \\
Oncology         & 88.00\% & 100.00\% & 81.50\% & 91.50\%  \\
Ophthalmology    & 44.00\% & 97.50\%  & 73.50\% & 97.00\%  \\
Orthopedics      & 81.50\% & 99.00\%  & 84.00\% & 99.00\%  \\
Otolaryngology   & 60.50\% & 100.00\% & 90.00\% & 99.50\%  \\
Pediatrics       & 78.50\% & 100.00\% & 85.50\% & 99.00\%  \\
Psychiatry       & 23.50\% & 99.50\% & 78.00\% & 97.50\%  \\
Respiratory      & 61.00\% & 97.00\%  & 63.50\% & 92.00\%  \\
Urology          & 59.00\% & 100.00\% & 64.50\% & 95.50\%  \\ \hline
\textbf{Overall}  & \textbf{66.14\%}  & \textbf{98.76\%} & \textbf{76.98\%} & \textbf{95.31\%}  \\ \hline
\end{tabular}
\end{table}

Table~\ref{tab: all department acc before and after evolution internal} presents the performance changes of doctor agents across all clinical departments before and after evolving in treating 20,000 patients. First, significant improvements are observed across all clinical departments, with maximum improvements of 76\% on the examination task and 31\% on the diagnosis task. Second, the overall accuracy increases by 32.62\% for the examination task and 18.33\% for the diagnosis task, demonstrating that the proposed framework is effective and generalizable across all clinical departments in medical scenarios. Third, the smaller improvement observed for the diagnosis task compared to the examination task highlights the greater complexity and difficulty of diagnosis tasks, consistent with real-world clinical challenges. The averaged performance of evolved agents is all higher than 95\%, showing the effectiveness of our proposed model.

\subsection*{C.5 Main Experimental Results on the MedQA Dataset}
In Table~\ref{tab:all_results_on_MedQA}, we summarize the performance of different methods with distinct foundation models. First, \textit{MedAgent-Zero} outperforms state-of-the-art methods on all base models, though there are no labeled data used in \textit{MedAgent-Zero}. Second, \textit{MedAgent-Zero} with a basic foundation model can outperform a better foundation model, since the accuracy of \textit{MedAgent-Zero} on GPT-4 outperforms the Direct method on GPT-4o by 1.49\%. Third, the accuracy can be improved further when real-world data is added to our framework, as the accuracies of \textit{MedAgent-Zero (Hybrid)} are higher than \textit{MedAgent-Zero} in most settings.

\begin{table}[!h]

    \caption{Experimental results of different methods with distinct foundation models on the MedQA dataset. \textit{MedAgent-Zero (Hybrid)} means MedQA's training data is also added to the medical case base.} 
\label{tab:all_results_on_MedQA}
\centering
\begin{tabular}{lrrrr}
\hline
\textbf{Methods} & \textbf{GPT-3.5} & \textbf{GPT-4} & \textbf{GPT-4o} & \textbf{o1-preview}  \\ \hline
Direct     & 58.29 & 78.16 & 88.22 & 95.05  \\
CoT        & 64.02 & 83.11 & 90.42 & -                                 \\
MedAgents  & 66.30 & 84.45 & 89.24 & -                                  \\
Medprompt* & 71.09 & 88.30 & 91.12 & 94.50                              \\
Medprompt  & 73.76 & 89.47 & 91.52 & {\ul 95.36}                       \\
\textit{MedAgent-Zero}    & {\ul 74.31}      & {\ul 89.71}    & {\ul 92.22}     & \textbf{96.15}                        \\
\textit{MedAgent-Zero (Hybrid)}  & \textbf{76.83}   & \textbf{91.20} & \textbf{92.77}  & \textbf{96.15}                          \\ \hline
\end{tabular}
\end{table}

\begin{figure}[htbp]
    \centering
    \includegraphics[width=0.95\linewidth]{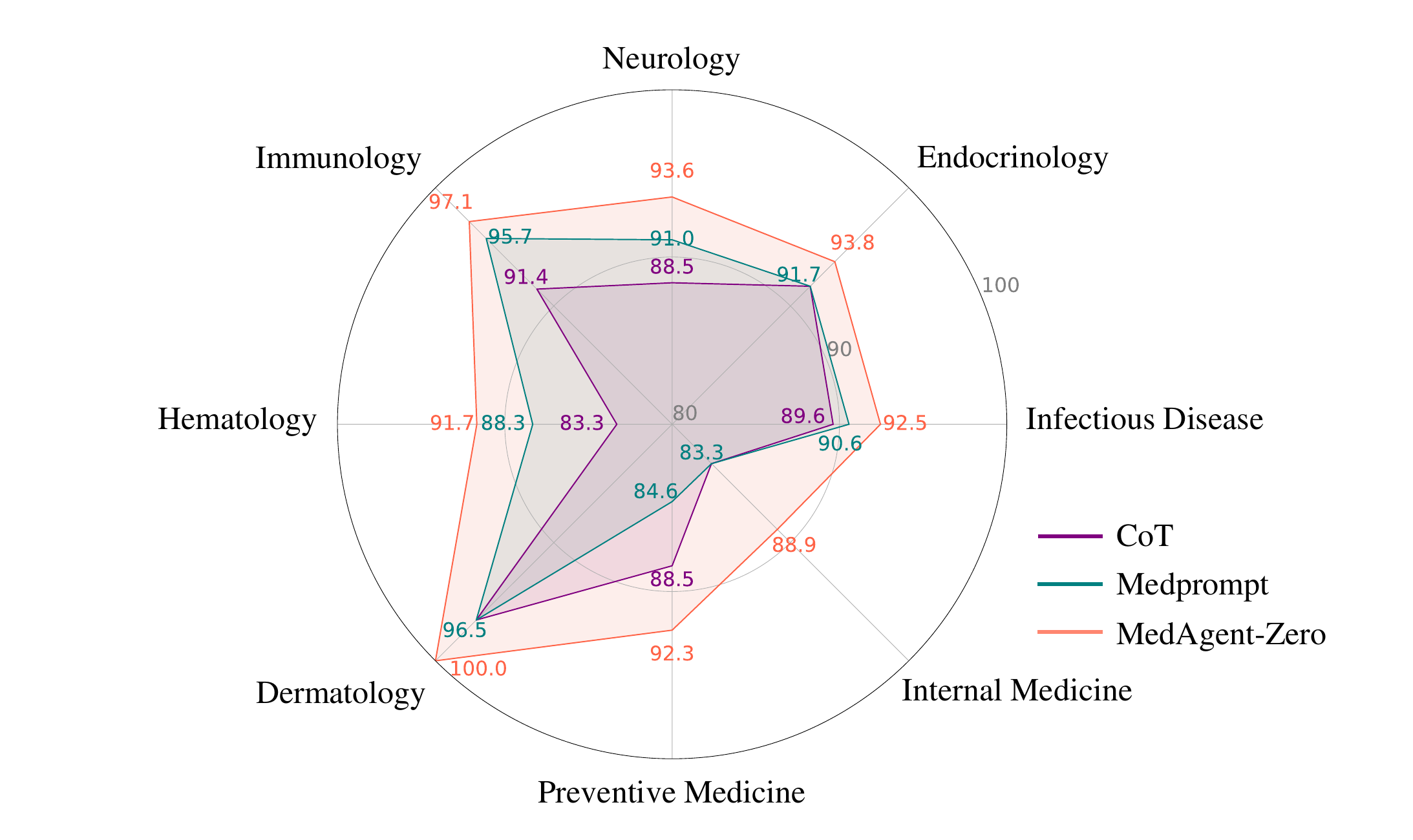}
    \caption{Comparison of CoT, Medprompt, and MedAgent-Zero across 8 clinical and non-clinical departments on the MedQA dataset. The reported values indicate the accuracy in the department.}
    \label{fig:radar}
\end{figure}

Some further department-level comparisons are shown in Figure~\ref{fig:radar}. \textit{MedAgent-Zero} consistently outperforms the other two methods, and Medprompt outperforms the CoT model in most departments.

\subsection*{C.6 Hyperparameter Analysis on the MedQA Dataset}
\begin{figure}[htbp]
    \centering
    \includegraphics[width=\linewidth]{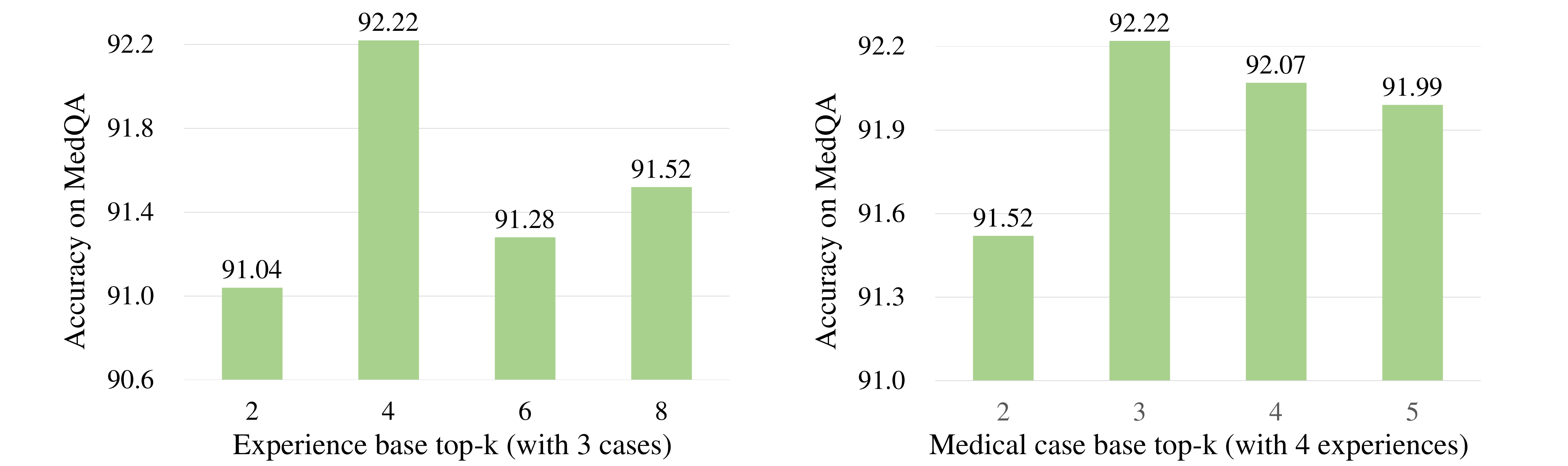}
    \caption{Hyperparameter analysis of top-k experiences and medical cases. During the adjustment of top-k experiences from the experience base, the top-3 medical cases from the medical case base are fixed. Similarly, when adjusting the top-k medical cases from the medical case base, the top-4 experiences from the experience base are kept constant. The reported values represent model accuracy on the whole MedQA dataset.}
    \label{fig:hyper-parameter}
\end{figure}

To evaluate the influence of top-k experiences and top-k medical cases during the inference phase in MedQA Dataset, we conduct experiments using various combinations of these two hyperparameters. After conducting pilot experiments, the top-4 experiences were fixed when adjusting the top-k medical cases from the medical case base, and the top-3 medical cases were fixed when adjusting the top-k experiences from the experience base. The results are presented in Figure~\ref{fig:hyper-parameter}. 

Experimental results reveal that when the top-4 experiences are fixed, the best performance is achieved at top-3 medical cases during inference. Increasing the number of medical cases beyond this point degrades performance, likely due to the inclusion of irrelevant information that disrupts the doctor agent’s reasoning process. Conversely, using fewer medical cases also reduces performance, possibly due to insufficient information for accurate decision-making. Similarly, when the top-3 medical cases are fixed, deviations from the optimal top-4 experiences—either increasing or decreasing the number—lead to worse performance. While a slight improvement is observed as the number of experiences increases from 6 to 8, the performance gap between top-8 experiences and the optimal top-4 experiences remains significant.

\begin{figure}[htbp]
    \centering
    \includegraphics[width=0.78\linewidth]{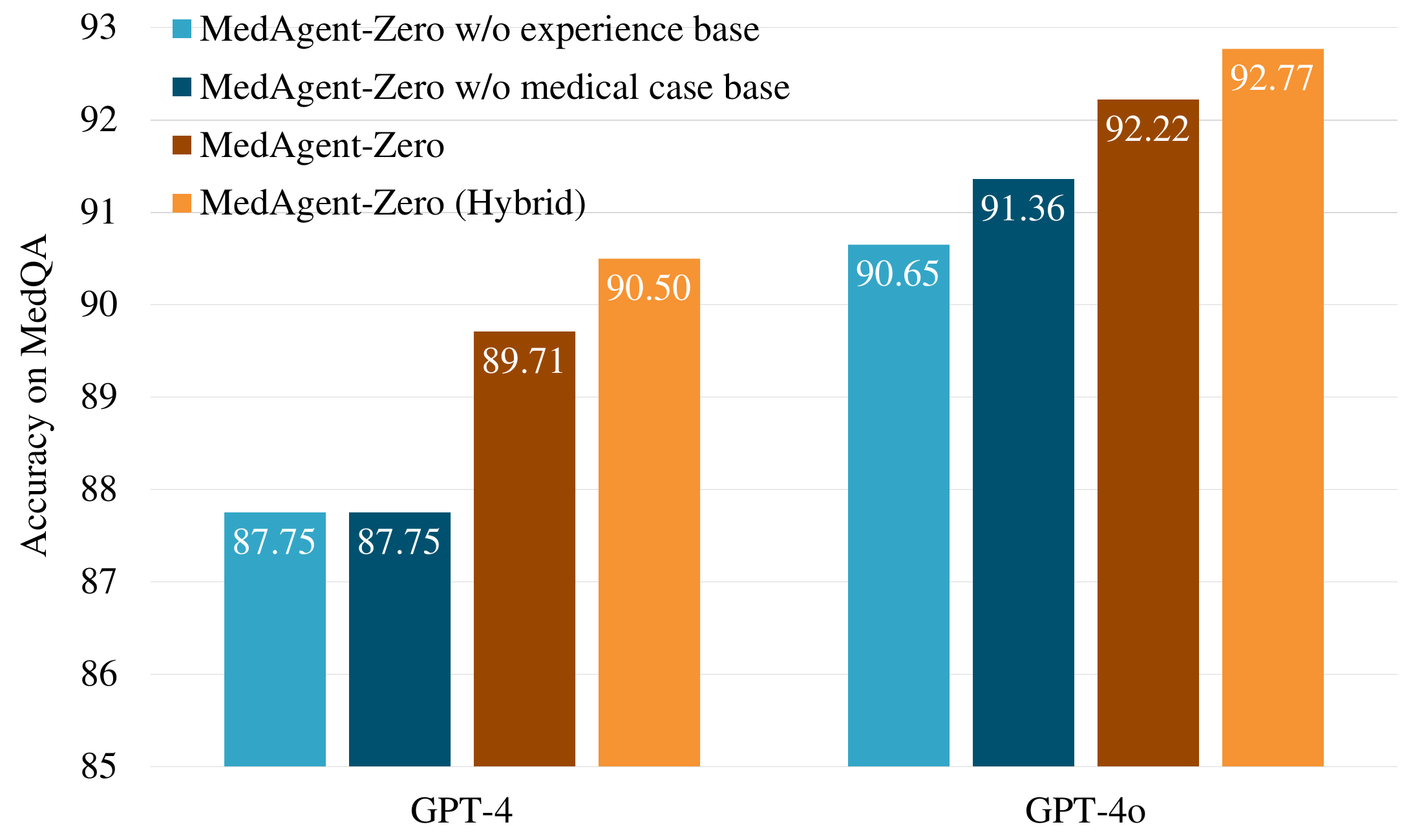}
    \caption{Ablation studies of \textit{MedAgent-Zero} on the MedQA dataset. \textit{MedAgent-Zero (Hybrid)} refers to incorporating the real-world Q\&A pairs into the respective medical case base of the doctor agents.}
    \label{fig:case_comparison}
\end{figure}

\subsection*{C.7 Ablation Study of MedAgent-Zero on the MedQA Dataset}
To further validate the effectiveness of the proposed medical case base and experience base, we conduct an ablation study of MedAgent-Zero on the MedQA Dataset, which is shown in Figure~\ref{fig:case_comparison} to represent the accuracy on the MedQA dataset. First, \textit{MedAgent-Zero}, which utilizes both the medical case base and experience base, achieves superior performance compared to using either records or experience alone, showing that both the experience base and medical case base are helpful. The results demonstrate the synergistic effect of the two components and indicate the experience base may be more helpful. Second, the inclusion of real-world data into \textit{MedAgent-Zero} further enhances performance, underscoring the benefit of combining virtual and real-world data for improved effectiveness in real-world tasks.

\begin{figure}
    \centering
    \includegraphics[width=0.95\linewidth]{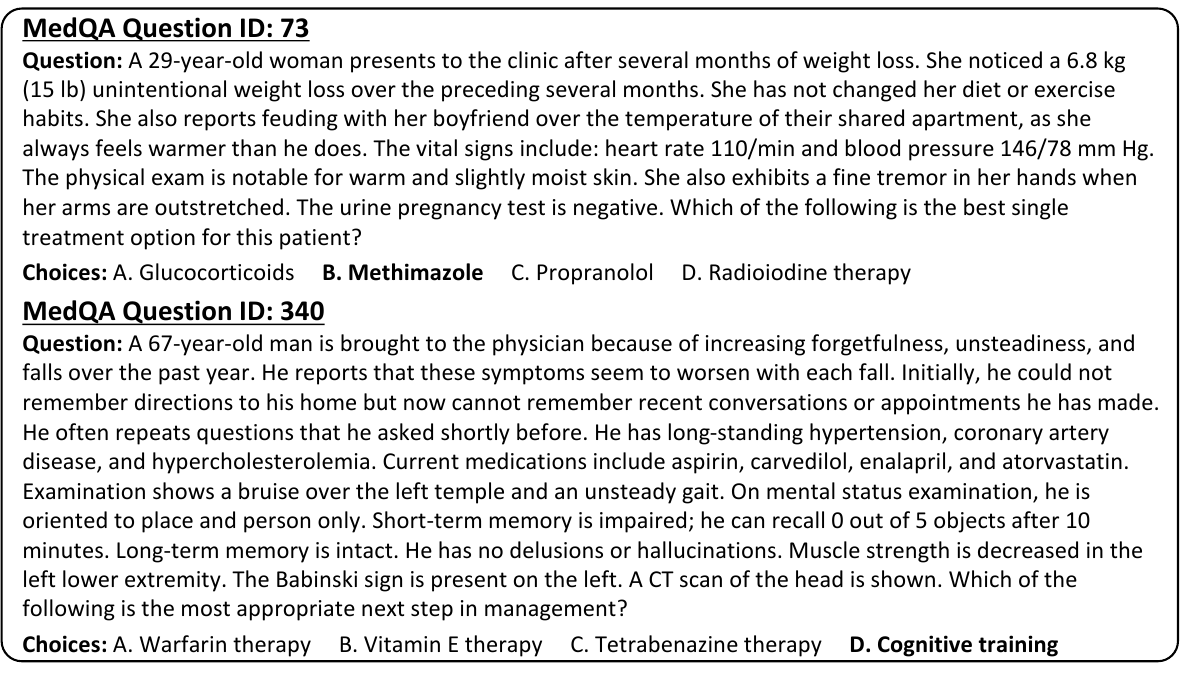}
    \caption{The content of the two MedQA questions, and the bolder choice is the correct answer.}
    \label{fig:questions}
\end{figure}

\section*{D. Case Studies}
To further demonstrate the effectiveness of the \textit{MedAgent-Zero} framework, case studies are conducted on two distinct questions from the MedQA dataset (IDs 73 and 340). The details of the medical questions can be found in Figure~\ref{fig:questions}, and all models in this section are driven by GPT-4o.

\subsection*{Analysis of Question 73}
\begin{figure}
    \centering
    \includegraphics[width=\linewidth]{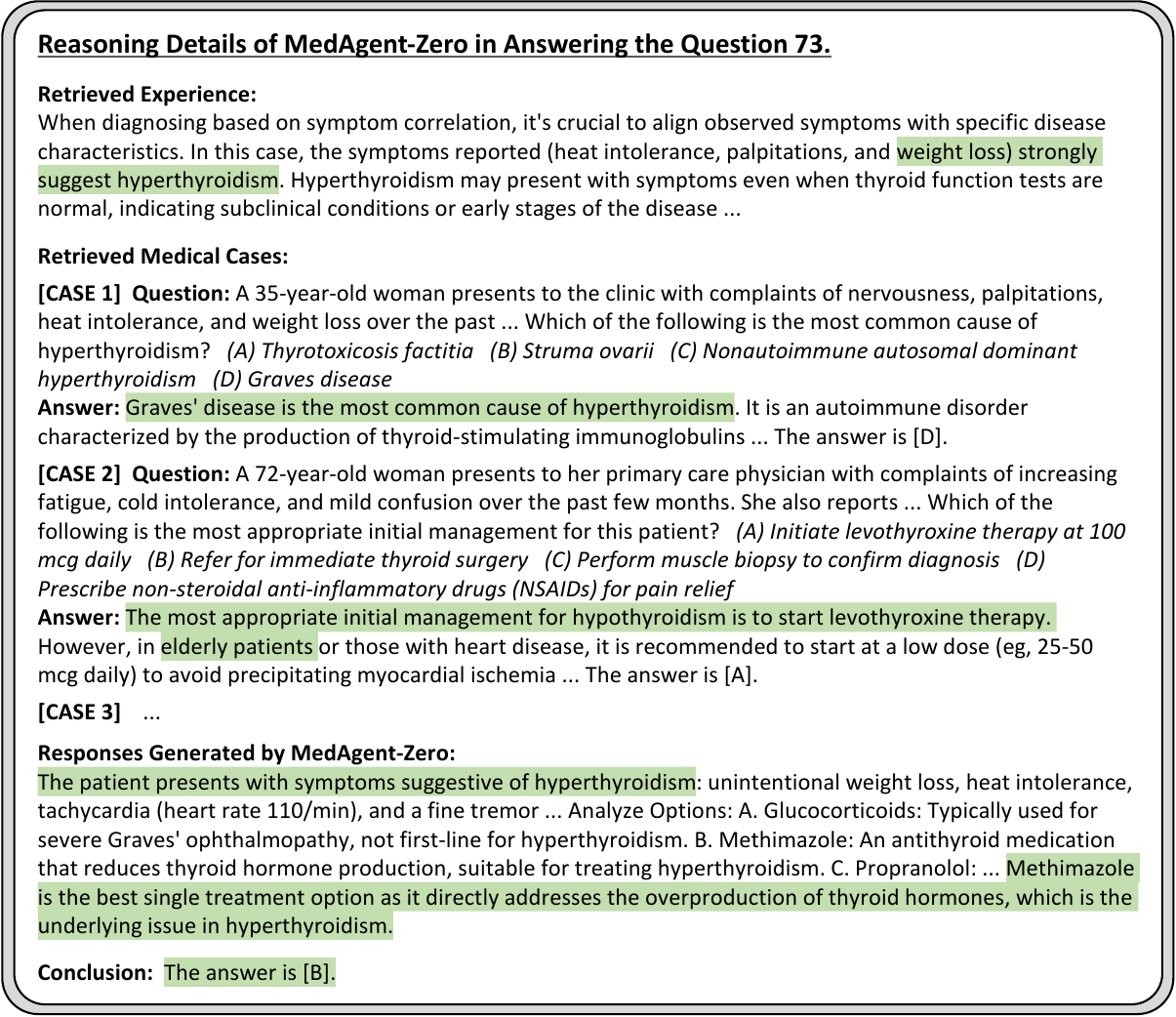}
    \caption{Reasoning details of \textit{MedAgent-Zero} in answering question 73. The retrieved medical experience and medical cases are both helpful. The green-highlighted text shows the usefulness of experiences/reasoning that contribute to the final correct answer.}
\label{fig:case2_medagent_zero}
\end{figure}

\begin{figure}
    \centering
\includegraphics[width=\linewidth]{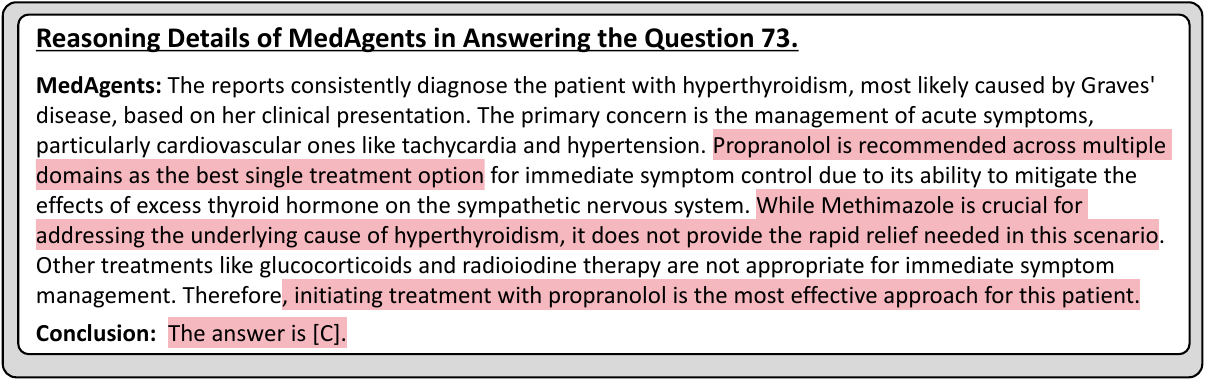}
    \caption{Reasoning details of MedAgents in answering question 73. The red-highlighted text reveals an error where the framework correctly identifies the disease as hyperthyroidism but fails to recommend the appropriate therapeutic plan. While propranolol (Option C) effectively manages acute symptoms like tachycardia and hypertension, the correct answer, methimazole (Option B), addresses the underlying cause of hyperthyroidism. The framework overlooks key considerations such as the long-term effectiveness of medications, the urgency of symptom control, and the patient's age. This example highlights the need for improved reasoning in balancing immediate symptom relief with definitive treatment to ensure clinically appropriate recommendations.}
    \label{fig:case2_medagents}
\end{figure}

\begin{figure}
    \centering
    \includegraphics[width=\linewidth]{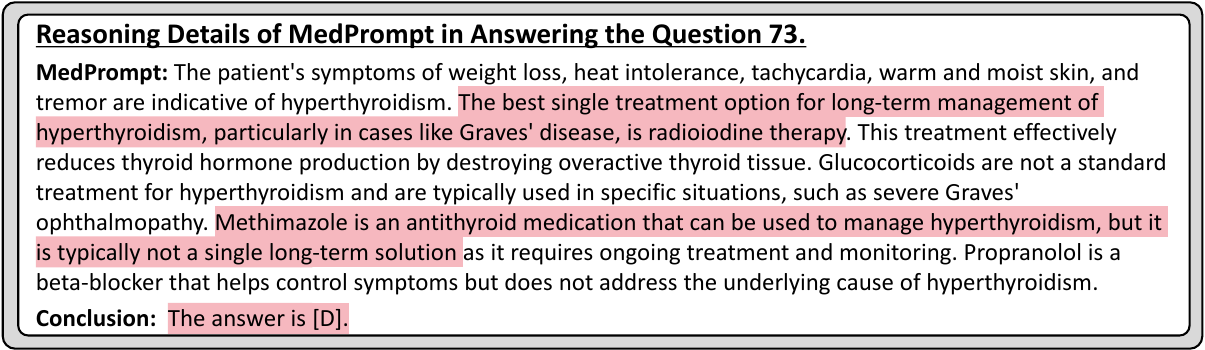}
    \caption{Reasoning details of MedPrompt in answering question 73. The red-highlighted text reveals an error where the framework correctly identifies the disease as hyperthyroidism but fails to recommend an appropriate therapeutic plan. While radioiodine therapy is effective for long-term management, it is not the best single treatment option in this scenario given the patient's acute symptoms, such as tachycardia and tremor. Methimazole (Option B), an antithyroid medication, is the most suitable choice for addressing the underlying cause while providing comprehensive management. The framework overlooks critical factors, including the urgency of symptom control, the patient's age, and the appropriateness of treatments within the context of acute versus long-term management. This example highlights the need for improved reasoning in aligning therapeutic recommendations with clinical priorities.}
    \label{fig:case2_medprompt}
\end{figure}

We show the correct reasoning of \textit{MedAgent-Zero} in Figure~\ref{fig:case2_medagent_zero}, and MedAgents and MedPrompt provide incorrect answers in Figures~\ref{fig:case2_medagents} and~\ref{fig:case2_medprompt}. Although both baselines correctly identify the disease as hyperthyroidism, they fail when recommending therapeutic plans, which requires consideration of factors such as the effectiveness of medications, the urgency of symptoms, and the patient's age and gender.

MedAgents recommends Propranolol, which alleviates the patient’s symptoms. However, since Propranolol does not address the root cause of hyperthyroidism, this recommendation is deemed incorrect. MedPrompt, on the other hand, identifies that Propranolol is insufficient for treating the underlying disease and instead recommends Radioiodine therapy, a treatment that effectively reduces thyroid hormone production by targeting overactive thyroid tissue. Nevertheless, considering the patient’s age and gender, Radioiodine therapy is not the optimal therapeutic choice for this particular case.

In contrast, as shown in Figure~\ref{fig:case2_medagent_zero}, the doctor agent within the \textit{MedAgent-Zero} framework correctly recommends Methimazole. This is the most appropriate treatment as it targets the root cause of hyperthyroidism while minimizing side effects for the patient. The recalled experience and medical cases play a crucial role in enabling the doctor agent to accurately diagnose the patient’s condition. Specifically, the recalled experiences support the doctor agent in identifying the patient's hyperthyroidism, while the first recalled case provides critical insight into its primary cause, Graves’ disease. Additionally, the second recalled case highlights the importance of considering the patient’s age, which proves essential in excluding Radioiodine therapy as an unsuitable treatment option.

\subsection*{Analysis of Question 340}
\begin{figure}
    \centering
    \includegraphics[width=\linewidth]{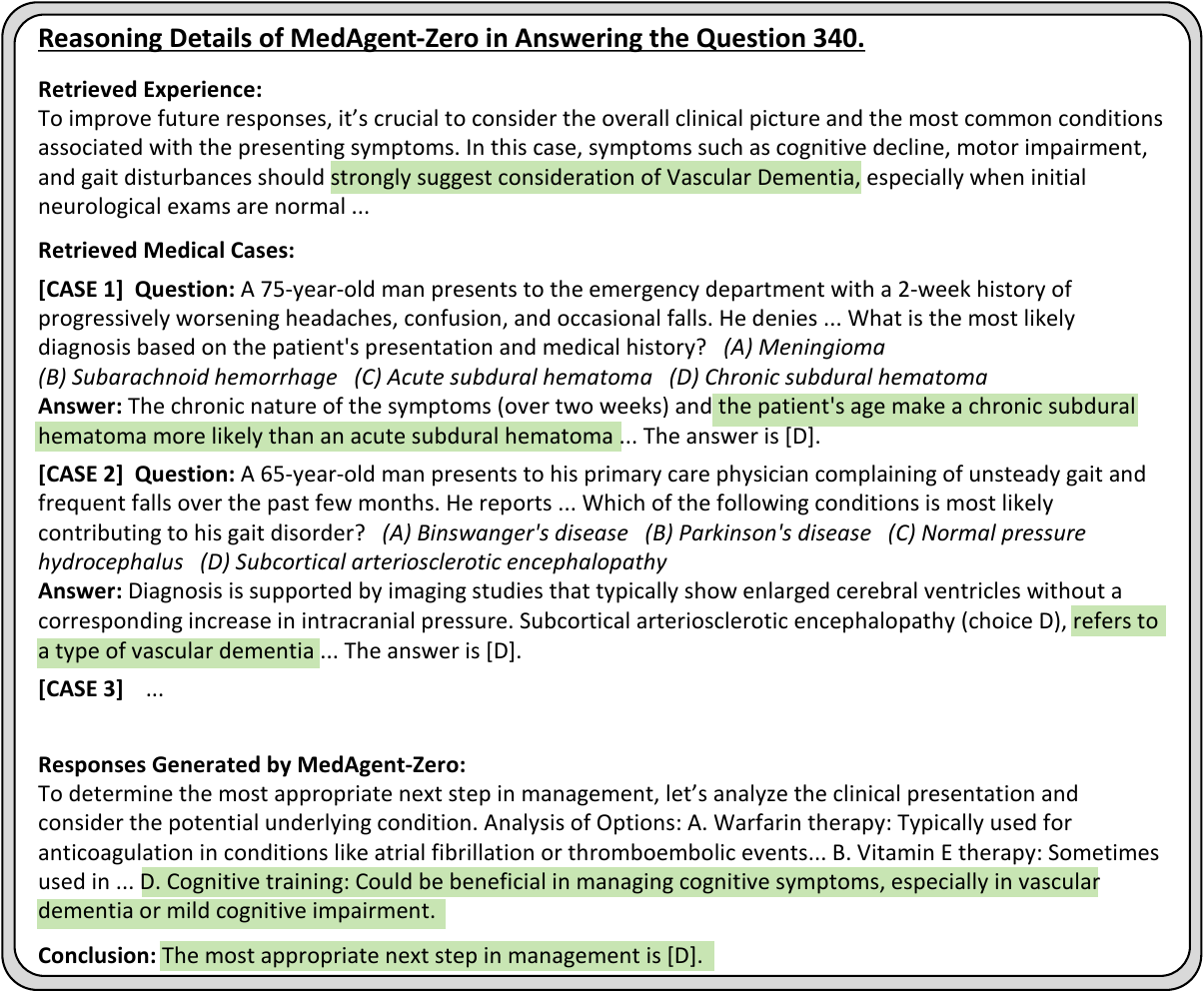}
    \caption{Reasoning details of \textit{MedAgent-Zero} in answering question 340. The retrieved medical experience and medical cases are both helpful. The green-highlighted text shows the usefulness of experiences/reasoning that contribute to the final correct answer.}
    \label{fig:case1_medagent_zero}
\end{figure}

\begin{figure}
    \centering
    \includegraphics[width=\linewidth]{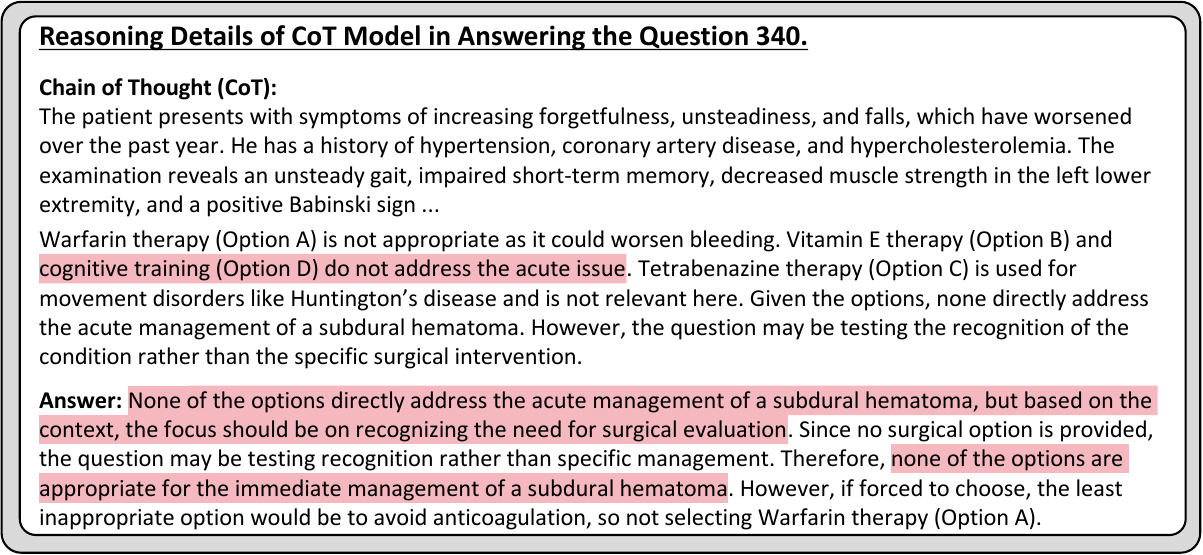}
    \caption{Reasoning details of the CoT model in answering question 340. The red-highlighted text indicates a critical error in analysis, where the framework incorrectly identifies the condition as a subdural hematoma. This misdiagnosis leads to a flawed interpretation of the question's requirements and the available options. Despite the incorrect diagnosis, the model appropriately identifies Warfarin therapy (Option A) as the least inappropriate choice, based on the provided options and the need to avoid anticoagulation. The example highlights a limitation in CoT reasoning when confronted with cases requiring recognition of surgical intervention rather than direct management from the given options.}
    \label{fig:case1_CoT}
\end{figure}

\begin{figure}
    \centering
    \includegraphics[width=\linewidth]{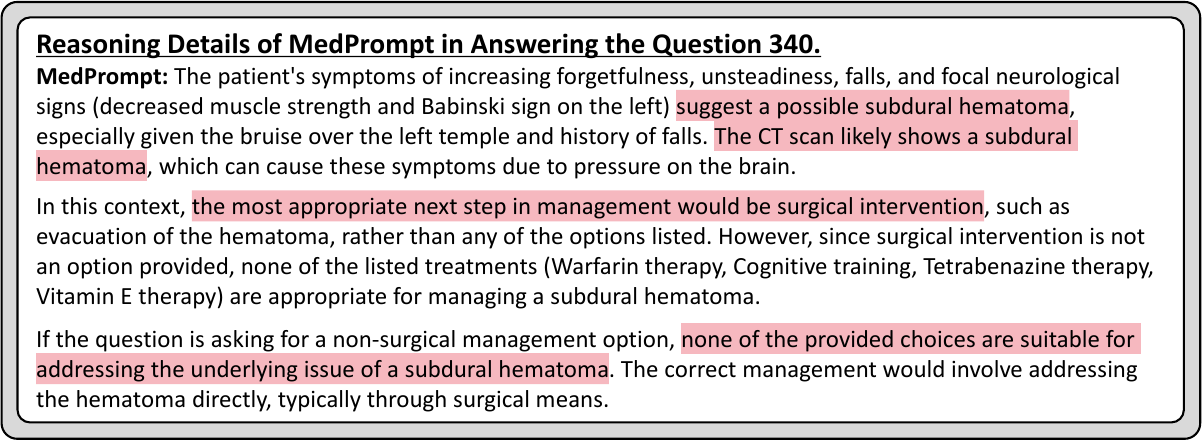}
    \caption{Reasoning details of MedPrompt in answering question 340. The red-highlighted text illustrates a reasoning error where the framework incorrectly diagnoses the condition as a subdural hematoma based on symptoms such as unsteadiness, memory loss, and a CT scan interpretation. This misdiagnosis leads to the recommendation of surgical intervention, which is not among the provided options. Consequently, the method fails to identify the correct answer, as none of the listed options (Warfarin therapy, Cognitive training, Tetrabenazine therapy, or Vitamin E therapy) align with the misinterpreted diagnosis. This example highlights a key limitation in MedPrompt’s reasoning when no options correspond to the misdiagnosed condition.}
    \label{fig:case1_medprompt}
\end{figure}

We show the correct reasoning of \textit{MedAgent-Zero} in Figure~\ref{fig:case1_medagent_zero}.
The outputs from the CoT model and MedPrompt are also listed in Figures~\ref{fig:case1_CoT} and~\ref{fig:case1_medprompt}. Both CoT and MedPrompt incorrectly diagnose the condition as subdural hematoma and recommend surgical intervention. Consequently, neither method identifies the correct answer from the provided options, as no relevant choice aligns with this misdiagnosis. 

In contrast, Our \textit{MedAgent-Zero} framework accurately recognizes the patient’s condition as vascular dementia rather than subdural hematoma, leading to the correct answer: cognitive training, which aids in the patient’s recovery. This improved performance is attributed to \textit{MedAgent-Zero}'s ability to enable the doctor agent to reference similar patient cases and recall relevant experiences when addressing the current question. 
As illustrated in Figure~\ref{fig:case1_medagent_zero}, the doctor agent retrieves experiences related to diagnosing vascular dementia and medical cases associated with both vascular dementia and subdural hematoma. By considering the patient’s age and symptoms alongside these retrieved experiences and medical cases, the doctor agent effectively determines that the patient’s condition is vascular dementia. This case highlights the advantage of leveraging prior experiences and case-based reasoning in \textit{MedAgent-Zero}.

Notably, all the experiences and medical cases retrieved by the \textit{MedAgent-Zero} originate from unlabeled data, demonstrating \textit{MedAgent-Zero}’s remarkable capability.

\end{document}